# Vision Transformers for Zero-Shot Clustering of Animal Images: A Comparative Benchmarking Study

Hugo Markoff[1] | Stefan Hein Bengtson[2] | Michael Ørsted[1]

[1]Department of Chemistry and Bioscience, Aalborg University, Aalborg, Denmark | [2]Department of Architecture, Design and Media Technology, Visual Analysis and Perception Lab, Aalborg University, Aalborg, Denmark

**Correspondence:** Hugo Markoff (khbm@bio.aau.dk)

## ABSTRACT

Manual labelling of animal images remains a significant bottleneck in ecological research, limiting the scale and efficiency of biodiversity monitoring efforts. This study investigates whether state-of-the-art Vision Transformer (ViT) foundation models can reduce thousands of unlabelled animal images directly to species-level clusters. We present a comprehensive benchmarking framework evaluating five ViT models combined with five dimensionality reduction techniques and four clustering algorithms, two supervised and two unsupervised, across 60 species (30 mammals and 30 birds), with each test using a random subset of 200 validated images per species. We investigate when clustering succeeds at species-level, where it fails, and whether clustering within the species-level reveals ecologically meaningful patterns such as sex, age, or phenotypic variation. Our primary, fully unsupervised finding is that DINOv3 embeddings with t-SNE and HDBSCAN reach a species-level V-measure of 0.943 with no prior knowledge of the number of species, rejecting only 1.14% of images as outliers requiring expert review. Assigning one label per cluster reduces manual labelling actions by 98.3%, with 94.6% of clustered images labelled correctly, close to the accuracy of manual annotation. When the true number of species is supplied, supervised hierarchical clustering raises this to 0.958, which we report as an upper bound as the species count is typically unknown in real deployments. Benchmark numbers are reported for the 2D space used throughout, and we discuss practical deployment, including the effect of retaining more embedding dimensions for downstream analysis and where this changes the recommended method to UMAP. We further demonstrate robustness to realistic long-tailed distributions of species and show that intentional over-clustering can reveal intra-specific variation including age classes, sexual dimorphism, and pelage differences. We introduce an open-source benchmarking toolkit and provide recommendations for ecologists to select appropriate methods for sorting their specific taxonomic groups and data.

## 1 | Introduction

Biodiversity monitoring has entered a critical phase where the volume of visual data generated through camera traps, citizen science platforms, and research expeditions exceeds our capacity for manual analysis Weinstein [1]. At the same time, global wildlife populations are plummeting, where recent estimates suggest that monitored vertebrate populations have declined by 73% in the past five decades, making efficient and accurate species monitoring more crucial than ever [2]. Traditional approaches to animal image classification rely heavily on expert knowledge and manual annotation, creating a significant bottleneck in ecological research workflows.

With individual studies generating hundreds of thousands to millions of images annually, the volume of data fundamentally limits the scale and timeliness of biodiversity assessments if labelling of the data has to be done manually Norouzzadeh et al.

[3]. Figure 1 illustrates the core objective of this work: transforming an unordered collection of animal images into coherent species-level clusters without manual annotation.

Current solutions often employ supervised learning approaches, such as Convolutional Neural Networks (CNNs), which are trained on large labelled datasets to classify species [4, 5]. Binary animal classification models for automatic camera systems, in different domains, terrestrial mammals, birds, fishes and insects manage to achieve high accuracy in detecting solely the presence of animals in images.

These detection models include MegaDetector Beery et al. [4] for terrestrial mammals and birds, Community Fish Detector (CFD) WildHackers [6] for fishes, and FlatBug Svenning et al. [7] for insects. These models show great potential for grounding the input for species-level classification models, by cropping out the area of interest, where an animal is present. There have already been classifiers developed with this pipeline, where MegaDetector is used to crop parts of the image where an animal is present, and passing the croppings to a species-level classification model, such as DeepFaune Rigoudy et al. [8] and SpeciesNet Gadot et al. [5].

While achieving high accuracy on well-represented species and domains, supervised deep learning perpetuates an annotation bottleneck: thousands of images must be manually labelled before models become useful Norouzzadeh et al. [3]. Furthermore, supervised classifiers are inherently limited to a fixed set of species present in the training data Vyskočil and Picek [9] and struggle with the typical long-tailed distribution of biodiversity, where a few common species dominate, while many rare or threatened taxa have insufficient training examples Pantazis et al. [10]. This is especially problematic as the rarer species are often important and the lack of the sufficient training data may cause them to be wrongly classified. Furthermore, supervised classifiers trained on limited datasets can overfit to specific domains, failing to generalise to new sites with different backgrounds, imaging conditions, or species [3, 11], limiting their real-world applicability. These constraints create a need for methods that can organise and interpret unlabelled animal imagery without extensive species-specific training.

Recent advances in Vision Transformer (ViT) foundation models, particularly those trained on large-scale biological datasets such as BioCLIP Stevens et al. [12] and BioCLIP 2 Gu et al. [13], have demonstrated zero-shot classification capabilities by learning visual representations from millions of species images. However, there exists a critical gap between demonstrated classification performance and practical deployment scenarios, such as usage in autonomous camera systems. For instance, BioCLIP 2 states a 54% species-level accuracy for the IDLE-OO-Camera-Traps dataset Gu et al. [13], meaning that for the remaining 46% of images, extensive manual validation is still required.

Lowe and Haurum [14] demonstrated the feasibility of zero-shot clustering for ecological images, showing that image embeddings from pretrained models, such as self-supervised ViT ones, can form well-defined clusters, even on unseen fine-grained biodiversity datasets such as species from iNaturalist.

## 1.1 | Our Contribution

In this paper, we approach the problem from a practical ecological perspective, moving beyond feasibility demonstrations to address real-world deployment needs. Rather than testing whether models can match individual images to predefined species labels, we examine whether learnt representations naturally separate images by species when no prior labels are available. We systematically benchmark combinations of foundation models, dimensionality reduction techniques, and clustering algorithms to provide results from different combinations of model and method selections. We also investigate when these methods fail and whether clustering failures reveal biologically meaningful patterns (sex dimorphism, age classes, seasonal fur changes) versus model limitations. This work aims to provide results to better understand when clustering can reduce annotation burden, when it reveals intra-species variation, and when manual validation remains essential.

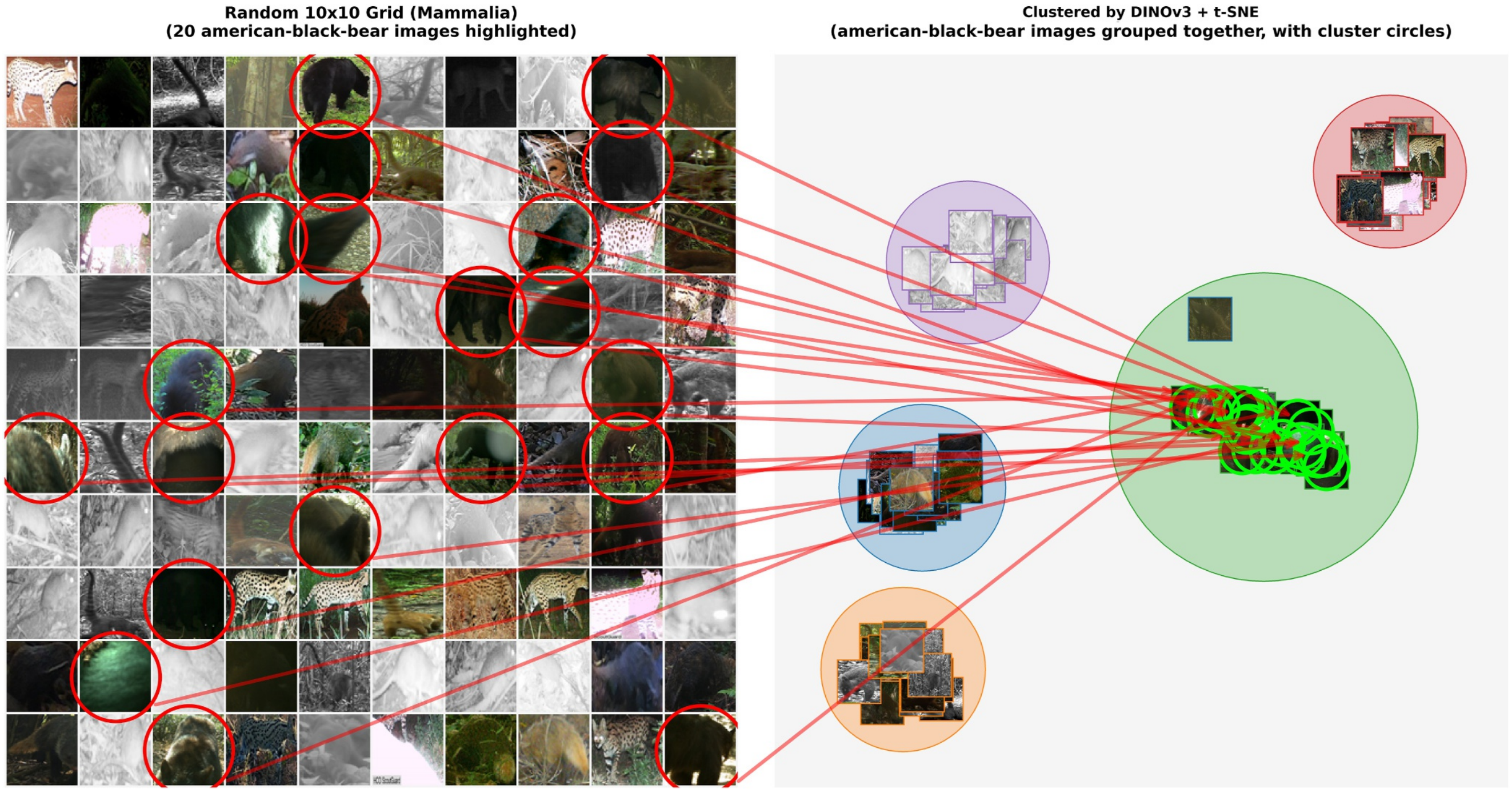

**FIGURE 1** | Left side, 100 randomly ordered images from 5 mammal classes. Right side, images sorted in 2-D embedding space clusters, from results of DINOv3, t-SNE and HDBSCAN. Images are from the camera trap datasets listed in Table 1.

Our work addresses these gaps through four primary contributions:

1. Comprehensive benchmarking framework: We systematically evaluate 5 state-of-the-art ViT models × 5 dimensionality reduction approaches × 4 clustering algorithms across a total of 60 species from two taxonomic classes (Mammals and Birds), providing quantitative performance metrics for thousands of method combinations and results to select best-performing combinations.
2. Validated animal dataset: We provide an open-source dataset consisting of 139,111 manually validated image crops across all 60 species, each processed through automated detection pipelines that mirror real-world workflows and enables benchmarking for future research.
3. Ecological interpretation framework: Beyond quantitative metrics, we analyse clustering compositions to distinguish when 'failures' represent genuine ecological subcategories (intra-species variation or environmental context with biological meaning) versus pure model limitations, providing insights for practical deployment and expert validation strategies.
4. Open-source toolkit and visualisation website: We release complete code, dataset, logged results from all experimental runs, and an interactive web interface, enabling ecologists to explore method combinations and visualise clustering results without coding expertise.
5. Practical deployment guidelines: Based on our empirical results, we provide concrete recommendations for parameter selection tailored to dataset characteristics, including guidance for large versus small datasets, handling long-tailed species distributions, and prioritising rare species detection for expert review.

Our approach explicitly targets the practical needs of ecological researchers, conservationists, land managers, and park rangers who handle large volumes of camera trap images, namely by reducing the annotation burden. Specifically, this is achieved by allowing review at the whole cluster level rather than one image at a time, handling taxonomic uncertainty, prioritising expert attention on ambiguous edge-cases, and understanding when automated methods can be trusted versus when manual validation remains essential.

## 2 | Related Work

### 2.1 | Vision Transformers in Computer Vision

Vision Transformers, introduced by Dosovitskiy et al. [15] revolutionised computer vision by adapting the transformer architecture from natural language processing to image analysis. By treating images as sequences of patches and applying self-attention mechanisms, ViTs achieved competitive performance with CNNs while offering better interpretability and scalability. The success of ViTs has led to numerous variants optimised for specific applications and computational constraints.

CLIP (Contrastive Language-Image Pre-training) Radford et al. [16] demonstrated the power of multimodal learning by jointly training vision and language models on large-scale internet data. This approach enabled zero-shot classification capabilities, where models could classify images based on natural language descriptions without task-specific training. SigLIP Zhai et al. [17] improved upon CLIP's training efficiency through sigmoid-based contrastive learning, achieving better performance with reduced computational requirements.

Self-supervised learning approaches like DINOv2 Oquab et al. [18] have shown capabilities in learning visual representations without manually-defined labels. Instead, these models rely on automatically generating pseudo-labels and learn from these in a supervised manner. These models excel at capturing semantic relationships between images, making them particularly suitable for clustering and similarity-based tasks.

### 2.2 | Biological Foundation Models

Stevens et al. [12] marked a significant advance by training on TreeOfLife-10M, a dataset containing images of over 454,000 taxa. The model achieved 16%–17% improvements over general-purpose CLIP in taxonomic classification tasks. Critically, BioCLIP demonstrated zero-shot capabilities: identifying species it had never seen during training by leveraging taxonomic relationships learnt from related species.

Gu et al. [13] scaled to TreeOfLife-200M and introduced hierarchical contrastive learning that models taxonomic structure. Beyond classification improvements (18.1% over BioCLIP), the model exhibited emergent properties as training scaled. They found that the internal feature embeddings of the model aligned with ecological and functional relationships; for example, freshwater and non-freshwater fish naturally separated in the embedding space of the model. Intra-species variations such as different life stages were also preserved within some subspaces.

Recent work by Gadot et al. [19] demonstrated that cropping to bounding boxes improves classification performance compared to whole-image analysis. This pre-processing step reduces background noise and focuses attention on animal features. Detection models like MegaDetector have achieved near-perfect accuracy for binary animal detection in camera trap images, making automated cropping pipelines practical for large-scale studies.

### 2.3 | Clustering and Dimension Reduction

Recent work by Lowe and Haurum [14] addressed zero-shot clustering using ViT embeddings for different datasets, including biological image data, demonstrating the feasibility of using zero-shot clustering methods.

Dimensionality reduction techniques such as UMAP McInnes et al. [20], t-SNE van der Maaten and Hinton [21], and PCA offer different trade-offs between preserving local structure, global relationships, and computational efficiency. UMAP has been adopted in single-cell transcriptomics Becht et al. [22] and other biological applications due to its computational efficiency and ability to preserve both local and global structure McInnes et al. [20]. In contrast, t-SNE excels at revealing local clustering patterns van der Maaten and Hinton [21] but can distort distances between distant clusters and is computationally more intensive than UMAP Kobak and Berens [23].

Clustering algorithm selection poses additional challenges. Density-based methods such as DBSCAN and HDBSCAN Campello et al. [24] can determine cluster numbers based on data density without requiring pre-specification of the number of estimated clusters. Hierarchical clustering provides interpretable dendrograms but exhibits computational complexity depending on linkage method and implementation Abboud et al. [25], limiting scalability to large datasets. Gaussian Mixture Models (GMM) Reynolds [26] handle overlapping clusters probabilistically by assuming specific distributional forms. While clustering algorithms have been evaluated on various types of embeddings, systematic evaluations establishing optimal combinations for deep learning embeddings in camera trap species identification remain limited.

## 3 | Methodology

Our methodology consists of two distinct pipelines: (1) dataset preparation, where we collect and manually validate camera trap images to establish ground truth labels, and (2) zero-shot clustering, where we systematically evaluate and benchmark combinations of ViT models, dimensionality reduction techniques, and clustering algorithms. This section describes both pipelines and our evaluation framework.

### 3.1 | Dataset Preparation Pipeline

Our dataset preparation pipeline, illustrated in Figure 2, established the ground truth labels for evaluating the zero-shot clustering methods. An initial total of 281,127 raw camera trap images from 60 different species were selected. We applied automated detection and cropping, followed by expert validation, resulting in 139,111 manually validated species-level crops spanning from 461 to 6831 images per species (Table A1, Appendix B).

#### 3.1.1 | *Species Selection and Image Extraction*

We selected the 60 species (30 mammal and 30 bird species) based on three criteria: (1) sufficient raw data (≥ 3000 candidate images across multiple projects), (2) representation of diverse geographic regions and ecosystems, and (3) varied morphological characteristics across species. All images were sourced from 23 camera trap projects within LILA BC (Labelled Information Library of Alexandria: Biology and Conservation) LILA [28], a repository containing over 10 million images from worldwide camera trap deployments. Projects selected for our dataset can be seen in Table 1.

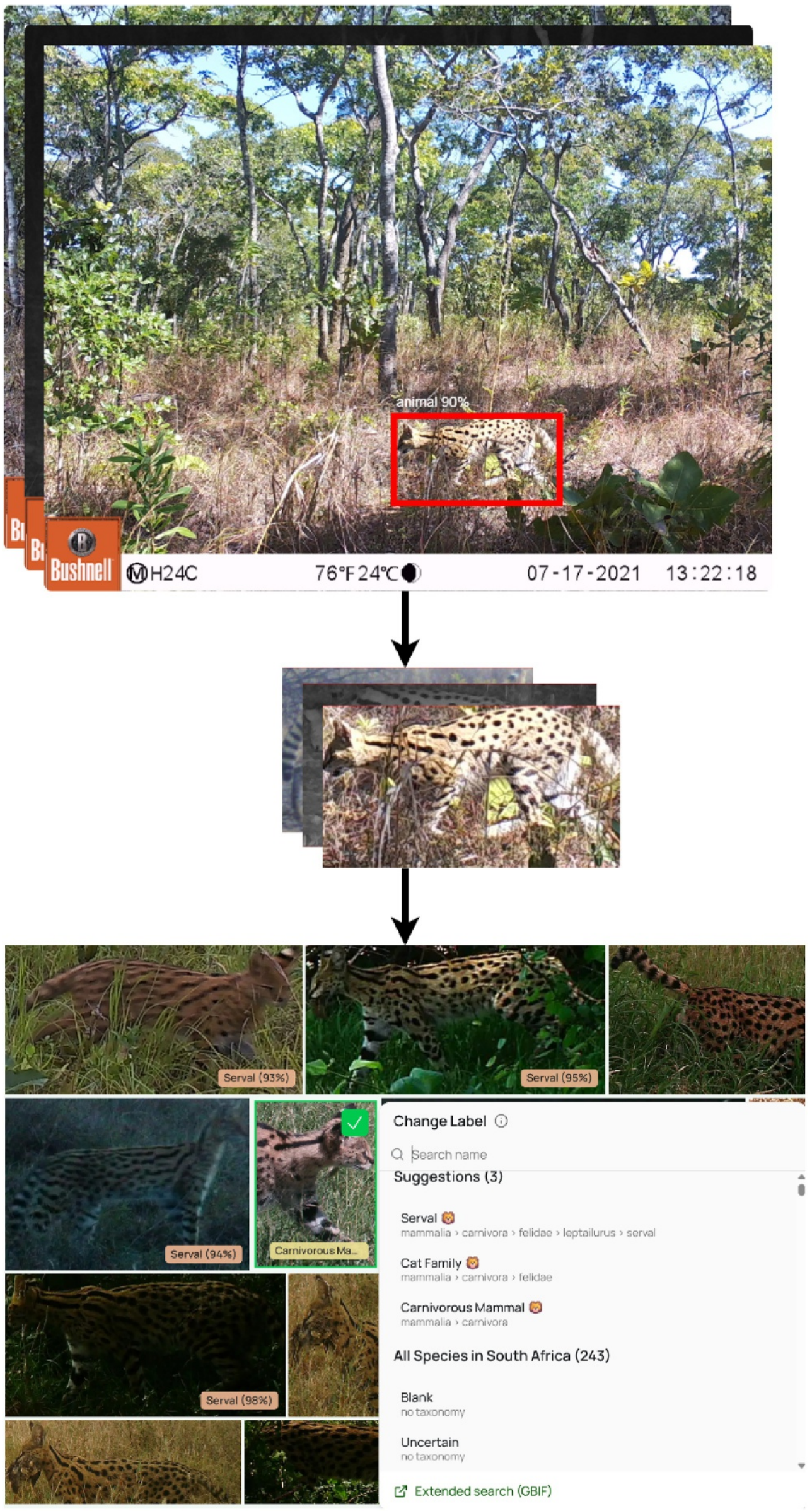

**FIGURE 2** | Data preparation pipeline using the Animal Detect platform. First step using the MegaDetector v1000 Redwood model to detect animals, shown with red bounding boxes. Second step, cropping the images, using the bounding box location and third step with expert validation, ensuring the classifications are correct, re-labelling uncertain cases. Serval images from the Nkhotakota Camera Traps dataset [27].

To handle taxonomic inconsistencies across projects, we developed a unification tool that standardized species labels within the 23 projects. For species with more than 5000 available images, the tool randomly sampled images, while species with fewer images included all available samples. With this multi-project sampling we attempt to create a heterogeneous dataset with images from different camera models, environmental backgrounds, lighting conditions, and animal poses. From the 60 selected species, we initially selected 281,127 raw images for processing. The complete species list, and the number of validated images, with sources is provided in Appendix B.

**TABLE 1** | Dataset source abbreviations.

| Code | Dataset |
|---|---|
| CAL | Caltech Camera Traps Beery et al. [29] |
| DLC | Desert Lion Camera Traps [30] |
| ENA | ENA24 Detection Yousif et al. [31] |
| ICT | Idaho Camera Traps Idaho Department of Fish and Game [32] |
| ISC | Island Conservation [33] |
| MIS | Missouri Camera Traps Zhang et al. [34] |
| NAC | NACTI Tabak et al. [35] |
| NKC | Nkhotakota Camera Traps Appel et al. [27] |
| NZC | New Zealand Camera Traps [36] |
| OSU | OSU Small Animals Balasubramaniam [37] |
| SCD | Snapshot Camdeboo [38] |
| SEN | Snapshot Enonkishu [39] |
| SIC | Seattle-ish Camera Traps Morris [40] |
| SKA | Snapshot Karoo [41] |
| SKG | Snapshot Kgalagadi [42] |
| SKR | Snapshot Kruger [43] |
| SMZ | Snapshot Mountain Zebra sna [44] |
| SS24 | Snapshot Safari 2024 [45] |
| SSE | Snapshot Serengeti Swanson et al. [46] |
| SWG | SWG Camera Traps SWG [47] |
| UNS | UNSW Predators [48] |
| WCS | WCS Camera Traps Wildlife Conservation Society [49] |
| WLC | Wellington Camera Traps Anton et al. [50] |

#### 3.1.2 | *Detection and Cropping*

Raw images were processed through the Animal Detect platform Markoff and Galaktionovs [51], where a simplified pipeline is illustrated in Figure 2, which implements:

Detection stage: MegaDetector models [4, 52, 53] have high recall for the detection of mammals and birds in camera trap images, allowing reliable automated processing. The MegaDetector v1000 Redwood model was selected for the accuracy gain reported over its predecessor MegaDetector v5a Morris [52]. The Microsoft release of MegaDetector v6 Microsoft AI for Good Lab [53] available at the time of testing did not reach the same performance on an internal test set (results not shown).

Classification stage: Detected regions are classified using SpeciesNet combined with Animal Detect's hierarchical reclassification logic Markoff and Galaktionovs [54], producing taxonomic predictions from kingdom to species-level with associated confidence scores.

Cropping: Images are cropped to bounding box coordinates following best practices established by Gadot et al. [19], where the cropped images are used in the zero-shot clustering pipeline, instead of the whole image.

Many images initially labelled with species annotations in LILA BC were filtered as empty during this stage. This occurred primarily due to camera burst mode, where a sequence of rapid-fire images receives the same species label even though the animal may only appear in one or two frames. Additionally, some images contained mislabelling in the original datasets. The pipeline intentionally retains challenging cases including partial occlusions, low-light conditions, motion blur, and images capturing only parts of animals (tails, legs, faces), reflecting real-world camera trap data quality.

Since the detection and cropping pipeline processes each bounding box independently, raw images containing multiple individuals of the same species would produce multiple crops.

#### 3.1.3 | *Expert Validation*

An ecologist with expertise in mammal and bird identification validated all the resulting image crops, verifying species-level identifications, while re-labelling very ambiguous cases as 'uncertain'.

This validation process resulted in a final dataset of 139,111 species-level crops, ranging from 461 to 6831 images per species. Of these, 138,024 images (99.2%) were confirmed at species-level, while 1087 images (0.8%) were marked as uncertain due to occlusion, blur, or ambiguous features. Both validated and uncertain images are included in the released dataset. The complete per-species breakdown of validated and uncertain images is provided in Table A1 (Appendix B).

### 3.2 | Zero-Shot Clustering Pipeline

The zero-shot clustering pipeline provides a modular framework for organising unlabelled animal images into clusters. The primary aim of the zero-shot pipeline is to cluster cropped animal images that have been through data preprocessing as explained in Section 3.1 into homogeneous clusters, representing species-level groups, without requiring labelled training data. The pipeline starts with cropped image detections and proceeds through three sequential stages (Figure 3), each of which can be configured with different methods depending on available resources and analytic goals.

Step 1: Feature extraction: Input cropped images of animals are passed through a pre-trained vision model to generate high-dimensional embedding vectors that capture visual semantics. The pipeline accepts models capable of producing fixed-length embeddings from images such as ViT models, hereunder different versions of DINO and CLIP, where the general purpose models would need no fine-tuning or task-specific training, enabling true zero-shot application.

Step 2: Dimensionality reduction (optional): High-dimensional embeddings can be projected to lower-dimensional spaces before clustering. This step serves two purposes: (1) potential improvement of clustering algorithm performance by focussing on relevant structure while reducing noise and computational complexity, and (2) enabling 2D/3D visualisation for manual inspection. The target dimensionality can be configured based on downstream requirements, from 2D for visualisation to higher dimensions that preserve more information. Clustering directly on original embeddings is also supported.

Step 3: Clustering: The embedding space is partitioned using clustering algorithms. *Unsupervised methods* automatically determine cluster count based on data geometry, suitable when species richness is unknown. *Supervised methods* require prespecification of cluster count, suitable when approximate species count is known.

#### *3.2.1 | Clustering Granularity and Extensibility*

The pipeline's output granularity depends on configuration choices and the underlying embedding quality. With appropriate settings, clusters could be used to approximate species-level groupings. However, the framework also supports finer-grained clustering that may reveal intra-specific variation such as sexual dimorphism, age classes, or phenotypic variants. This flexibility allows ecologists to tune the pipeline towards their specific analytical goals, from coarse taxonomic sorting to detailed population structure analysis. Higher cluster counts typically yield higher homogeneity (purer clusters) at the cost of splitting single species across multiple groups.

The modular design ensures the pipeline remains relevant as new methods emerge. Any component can be substituted without modifying other stages: new foundation models can replace the feature extractor, novel dimensionality reduction techniques can be incorporated, and alternative clustering algorithms can be evaluated. We release both the pipeline implementation and benchmarking code to facilitate extension with future methods (Section 6).

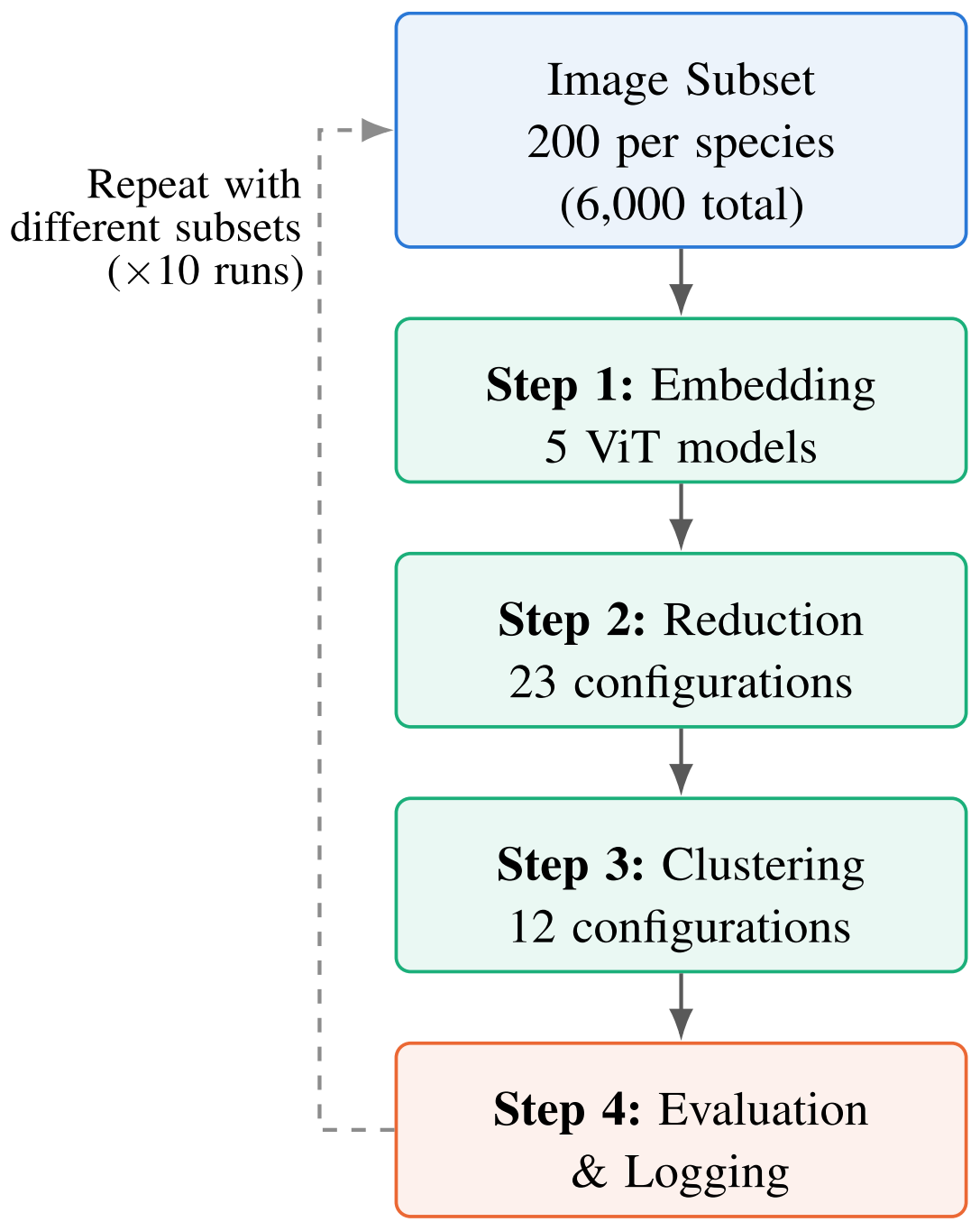

**FIGURE 3** | Benchmarking framework architecture. The complete pipeline is repeated 10 times for mammals and 10 for birds, with different random image subsets to generalise beyond a single specific sample selection.

### 3.3 | Benchmarking Framework

To evaluate the zero-shot clustering pipeline, we designed a benchmarking framework that systematically tests combinations of embedding models, dimensionality reduction methods, and clustering algorithms across multiple experimental runs.

#### *3.3.1 | Experimental Design*

The benchmarking framework seen in Figure 3 follows a factorial design, evaluating every combination of pipeline components to identify optimal configurations and understand component interactions.

Image Sampling: For each experimental run, we randomly sample 200 validated images per species within each taxonomic class (mammals or birds), yielding 6000 images per run (30 species × 200 images).

Experimental Scale: This benchmarking design evaluates:

$$\underbrace{10}_{\text{runs}} \times \underbrace{5}_{\text{ViT models}} \times \underbrace{23}_{\text{dim. red. configs}} \times \underbrace{12}_{\text{clustering configs}}$$
$$= 13,800 \text{ configs}$$

Additionally, clustering directly on original high-dimensional embeddings adds $10 \times 5 \times 12 = 600$ per class. The complete benchmark totals *28,800 configurations* across both taxonomic classes.

#### *3.3.2 | Vision Transformer Models*

We evaluate five state-of-the-art Vision Transformer models (Table 2) representing different training paradigms: general-purpose vision-language models (CLIP, SigLIP), self-supervised models (DINOv2, DINOv3), and a biology-specific model (BioCLIP 2). All models use publicly available pre-trained weights without fine-tuning, ensuring true zero-shot evaluation.

All sampled images are processed through each model to generate high-dimensional embedding vectors (768D–1536D depending on architecture). The embedding vector is extracted as the image representation, capturing global visual semantics.

**TABLE 2** | Vision transformer models evaluated.

| Model | Parameters | Embedding dim |
|---|---|---|
| CLIP (ViT-L/14)[a] | 304M | 768 |
| SigLIP (ViT-B/16)[b] | 200M | 768 |
| DINOv2 (ViT-g/14)[c] | 1100M | 1536 |
| DINOv3 (ViT-H+/16)[d] | 840M | 1280 |
| BioCLIP 2 (ViT-L/14)[e] | 304M | 768 |

[a]Radford et al. [16].
[b]Zhai et al. [17].
[c]Oquab et al. [18].
[d]Siméoni et al. [55].
[e]Gu et al. [13].

### 3.3.3 | *Dimensionality Reduction Techniques*

High-dimensional embeddings (768D–1536D) are projected to 2D spaces. While dimensionality reduction may improve clustering performance by reducing noise and computational complexity, the primary motivation for 2D projection is enabling visualisation for manual inspection for downstream evaluation tasks. We evaluate two non-linear methods alongside three deterministic. Each embedding set undergoes 23 reduction configurations:

Non-deterministic methods (10 runs each):

- UMAP (Uniform Manifold Approximation and Projection): Preserves both local and global structure through Riemannian geometry. Parameters: n_neighbours = 15, min_dist = 0.1, metric = 'euclidean', n_components = 2. Run 10 times with different random seeds to capture stochastic variation.
- t-SNE (t-distributed Stochastic Neighbour Embedding): Emphasises local neighbourhood structure through probability distributions. Parameters: perplexity = 30, learning_rate = 'auto', init = 'pca', n_components = 2. Run 10 times with different random seeds to capture stochastic variation. Unlike UMAP, t-SNE with the Barnes-Hut approximation, the default in common implementations and the variant used here, is in practice restricted to two and three output dimensions, because the size of its space-partitioning tree (a quadtree in 2D, an octree in 3D) grows exponentially with the output dimension van der Maaten [56]. The original exact solver supports any output dimension, but its $O(N^2)$ cost per iteration makes it impractical beyond small subsamples van der Maaten and Hinton [21].

Both manifold methods serve here as preprocessing for density-based clustering as well as for visualisation. t-SNE preserves local neighbourhoods but expands dense clusters, contracts sparse ones and does not preserve the distances between well-separated clusters [23, 57], so HDBSCAN and DBSCAN applied to its coordinates partly respond to the projection rather than to the geometry of the original embedding. We therefore treat t-SNE + HDBSCAN as a combined visualisation-and-clustering workflow. The primary benchmark uses 2D throughout. UMAP, by contrast, places no restriction on the output dimension McInnes et al. [20], so higher target dimensions are tested and discussed separately as a recommendation for workflows where more dimensions must be retained for downstream analysis.

Deterministic methods (1 run each):

- PCA (Principal Component Analysis): Linear projection capturing maximum variance. Provides interpretable baseline for non-linear methods.
- Isomap (Isometric Feature Mapping): Non-linear projection preserving geodesic distances along the data manifold.
- Kernel PCA: Non-linear projection using Radial Basis Function (RBF) kernel to capture complex relationships in feature space.

All embeddings are standardized using StandardScaler before reduction. Additionally, we evaluate clustering directly on original high-dimensional embeddings, with no reduction, bringing the total to 24 embedding space configurations per model.

### 3.3.4 | *Clustering Algorithms*

Each reduced embedding space is partitioned using four clustering algorithms representing two paradigms: unsupervised methods that autonomously discover cluster structure, and supervised methods requiring pre-specification of cluster count. This dual approach benchmarks both realistic field scenarios (unknown species counts) and informed scenarios (known approximate species richness).

Unsupervised Methods (2 configurations):

These algorithms automatically determine cluster count and structure based on embedding geometry, suitable for exploratory analyses where species count is unknown. Parameters were set using principled defaults rather than dataset-specific tuning to minimise bias and enable transfer to other ecological applications:

- DBSCAN (Density-Based Spatial Clustering): Groups points based on local density without requiring cluster count specification. Parameters: min_samples = 5, eps estimated automatically per dataset using the k-nearest neighbours heuristic. This automatic estimation adapts to the natural scale of each embedding space while reducing manual tuning requirements.
- HDBSCAN (Hierarchical DBSCAN): Builds a hierarchy of density-based clusters and extracts stable clusters at varying densities. Parameters: min_cluster_size = 15, min_samples = 5. The minimum cluster size of 15 reflects the practical consideration that smaller clusters are difficult to validate ecologically and may represent outliers rather than genuine species groupings.

Both density-based methods produce outlier labels $(-1)$ for points not meeting density criteria. In zero-shot clustering workflows, these outlier-labelled images require manual expert review, representing a trade-off between automated clustering and validation burden.

Supervised Methods (10 configurations):

These algorithms require pre-specification of cluster count, enabling systematic investigation of how incorrect species richness estimation affects performance. We evaluate multiple values: $k \in \{15, 30, 45, 90, 180\}$, where $k = 30$ represents the true species count. This range serves dual purposes:

1. Performance degradation quantification: Assess how clustering quality degrades when species richness estimates deviate from ground truth, spanning from half to six times the true count to capture estimation errors.
2. Ecological interpretation: Enable post-hoc analysis of cluster compositions. Under-splitting ($k < 30$) may reveal whether embeddings group species by higher taxonomic levels (e.g., families like cat family or deer family), while over-splitting ($k > 30$) may expose biologically meaningful intra-specific variation such as sexual dimorphism, age classes, seasonal pelage/plumage differences, or individual pattern variation.

Two complementary algorithms are evaluated at each $k$ value:

- Hierarchical Clustering: Agglomerative clustering with Ward linkage. Produces dendrograms, where distances between merge points indicate similarity. Evaluated at all five $k$ values.
- GMM (Gaussian Mixture Model): Probabilistic clustering assuming Gaussian component distributions. Parameters: covariance_type='full', random_state=42. Evaluated at all five $k$ values.

This combination of unsupervised (2 configs) and supervised (10 configs: 5 hierarchical + 5 GMM) methods yields 12 clustering configurations per reduced embedding space.

### 3.4 | Evaluation Metrics

During preliminary analysis, we computed several candidate metrics across all experimental configurations including Adjusted Rand Index (ARI) Hubert and Arabie [58], Adjusted Mutual Information (AMI) and Normalised Mutual Information (NMI) Vinh et al. [59], V-measure with its components homogeneity and completeness Rosenberg and Hirschberg [60], purity, and silhouette score Rousseeuw [61]. After analysing metric behaviour across the full experimental space, we selected two complementary metrics as our primary evaluation framework to compare results, while including more evaluation metrics in the supplementary results and appendixes. The two complementary metrics selected for primary results are:

V-measure: The harmonic mean of homogeneity and completeness Rosenberg and Hirschberg [60] (detailed formulations in Appendix A). V-measure provides interpretable decomposition: homogeneity measures whether clusters contain only one species, while completeness measures whether species are split across clusters. This directly addresses ecological concerns about false grouping versus species fragmentation. Range: 0 to 1, where higher values indicate better clustering.

Adjusted Mutual Information (AMI): Information-theoretic agreement between predicted and ground truth partitions, adjusted for chance Vinh et al. [59], see detailed formulations in Appendix A. AMI corrects for the bias that random clusterings achieve non-zero scores, making it critical when comparing methods with vastly different cluster counts, for example, HDBSCAN finding 35 clusters versus DBSCAN with 220 clusters. AMI serves as secondary validation, particularly for unsupervised methods. Range: 0 (random) to 1 (perfect), but can be negative when clustering performs worse than random.

Beyond quantitative metrics, we analyse cluster compositions to distinguish between true errors and biologically meaningful patterns. True errors include different species incorrectly grouped together (false positives) or same species incorrectly split without biological explanation (false negatives). However, some cluster splits may reveal fine-grained biological patterns which are manually inspected and documented. This analysis provides ecological value beyond pure species identification and informs when clustering can be trusted versus when expert validation remains essential.

The complete benchmarking code and the test results are released as open source (Section 6).

## 4 | Results

Our testing compare experimental configurations across five ViT models, five dimensionality reduction methods, and four clustering algorithms with varying parameters, resulting in a total of 27,600 combinations. The summarised results with detailed performance matrices is provided in Appendix C, while the key-findings are described in the following sections. The most promising configurations are used to further test and evaluate performance on unevenly distributed data and intra-species specific findings.

### 4.1 | Vision Transformer Model Performance

The selection of vision transformer model dramatically impacts clustering performance (Figure 4). Self-supervised DINO models substantially outperform the other vision-language models. DINOv3 achieves the highest average V-measure (0.817), followed by DINOv2 (0.769). In contrast, biology-specific BioCLIP 2 and general-purpose CLIP and SigLIP show substantially lower performance, with average V-measures ranging from 0.597 to 0.652. Notably, the distributions show consistent separation between models.

The performance advantage of DINO models extends across all dimensionality reduction and clustering method combinations. The complete result breakdown with additional performance metrics can be found in Appendix Table A5.

### 4.2 | Dimensionality Reduction

Across all experimental configurations, t-SNE and UMAP consistently outperformed PCA, Isomap and Kernel PCA by 26–38% points (Table 3, Figure 5). Not a single combination of PCA, Isomap or Kernel PCA achieved performance matching t-SNE or UMAP. Between the two best methods, t-SNE marginally outperforms UMAP by 0.6–1.1% points on average, with smaller differences for birds (0.6 pp) than mammals (1.1 pp). UMAP scores higher in 23% of individual configurations, particularly with density-based clustering methods, but is significantly ahead in only 4 of 60 model × method combinations (Section 4.4), as discussed in Section 5.2.

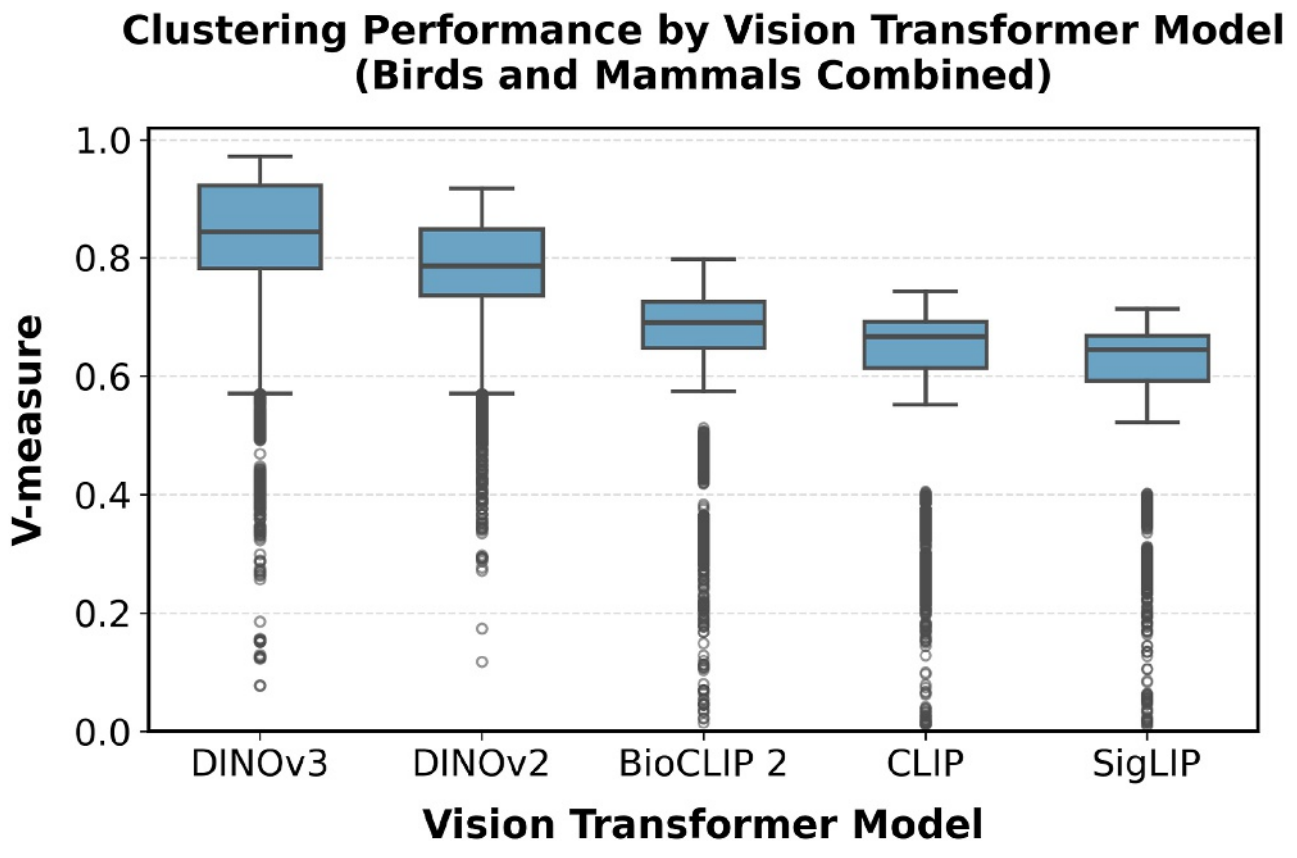

**FIGURE 4** | V-measure distribution across vision transformer models. Box plots show median, interquartile range, and outliers for all 5520 configurations per model.

### 4.3 | Supervised Clustering: Hierarchical and GMM Performance

When cluster count matches ground truth ($K = 30$), Hierarchical clustering and Gaussian Mixture Models (GMM) achieve nearly identical performance across all model × dimensionality reduction combinations. Hierarchical clustering averages marginally higher (+0.001 to +0.006 V-measure difference). The aggregate difference is consistent across subsamples but negligible in size, and for DINOv3 with t-SNE or UMAP it is not significant after correction (Table 4).

**TABLE 3** | Dimensionality reduction method comparison.

| Method | Birds | Mammals | Average | Versus t-SNE |
|---|---|---|---|---|
| t-SNE | **0.750** | **0.724** | **0.737** | — |
| UMAP | 0.744 | 0.713 | 0.729 | −0.8 pp |
| Isomap | 0.493 | 0.466 | 0.480 | −25.7 pp |
| PCA | 0.381 | 0.362 | 0.372 | −36.5 pp |
| KPCA | 0.382 | 0.325 | 0.354 | −38.3 pp |

*Note:* V-measure averaged across all experimental configurations for the different dimensionality reduction methods. The bold marks the best method (highest V-measure) in each column.
Abbreviation: pp, percentage points.

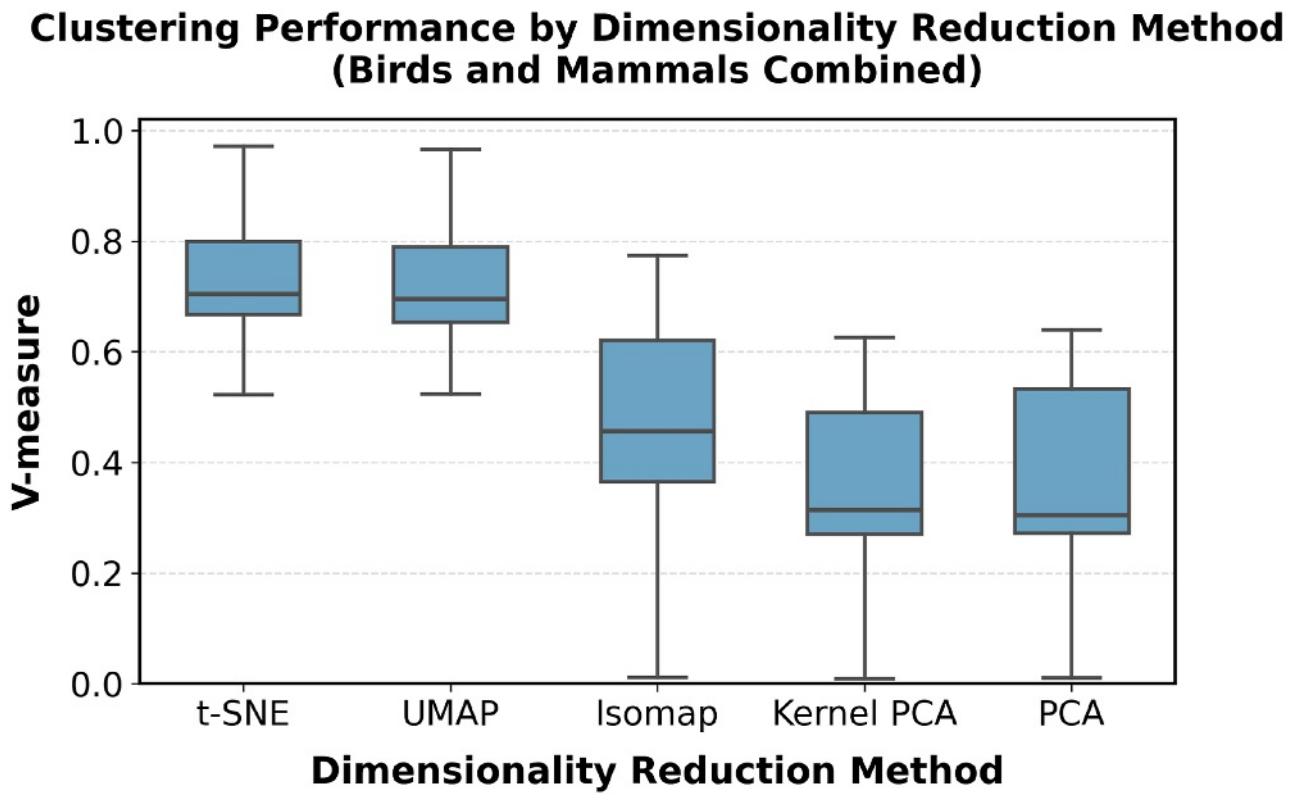

**FIGURE 5** | V-measure distribution across dimensionality reduction methods.

The choice of K affects performance, with distinct patterns emerging between model classes. For DINOv3 and DINOv2, optimal performance occurs at the true species count ($K = 30$), achieving average V-measures of 0.909 and 0.838 respectively in average across all the dimension reduction methods. These models maintain strong performance even under misestimation: at $K = 45$ (50% overestimate), DINOv3 retains 0.880 V-measure, representing only 3.2% degradation.

In contrast, the vision-language models BioCLIP 2, CLIP and SigLIP achieve better performance at $K = 45$ rather than the true $K = 30$. For these models, $K = 45$ configurations yield 1%–3% higher V-measures compared to $K = 30$. The AMI and V-measure across the different K values and across all five ViT models can be found in Appendix Table A3, where Hierarchical clustering and GMM differ by less than 0.01 at every K.

For DINOv3 combined with the top performing dimension reduction methods, t-SNE and UMAP, the average V-measure averages at 0.960 for birds and 0.952 for mammals, using Hierarchical clustering with $K = 30$ while GMM, $K = 30$ reaching an average V-measure of 0.957 for birds and 0.952 for mammals. See Appendix Table A2.

### 4.4 | Statistical Significance of the Rankings

To test whether these rankings are stable across the 10 random image subsets per taxonomic class, we ran paired tests over the 20 subsamples (taxonomic class × run). Each subsample contributes one value per method, its mean V-measure over all other pipeline settings, so that configurations sharing a subsample are not counted as independent. Table 4 reports the mean difference with a bootstrap 95% confidence interval (CI) Efron [62], an exact Wilcoxon signed-rank test Wilcoxon [63], and the number of model × method combinations in which each side is significantly ahead after Holm correction for multiple testing Holm [64]. All three aggregate differences are significant at $p < 10^{-5}$, but they differ greatly in size. The DINOv3 advantage over DINOv2 is large (0.048 V-measure) and holds in 34 of the 60 combinations, with DINOv2 significantly ahead in only 2. The t-SNE advantage over UMAP is significant but small (0.008), and UMAP is significantly ahead in 4 combinations, including DBSCAN with both DINO models. The Hierarchical

**TABLE 4** | Paired significance tests over the 20 image subsamples (10 Aves, 10 Mammalia).

| Comparison | Δ V-measure | 95% CI | Wins | Wilcoxon $p$ | Combinations ahead: First | Combinations ahead: Second |
|---|---|---|---|---|---|---|
| DINOv3 versus DINOv2 | 0.048 | [0.044, 0.052] | 20/20 | $< 10^{-5}$ | 34 of 60 | 2 of 60 |
| t-SNE versus UMAP | 0.008 | [0.007, 0.010] | 20/20 | $< 10^{-5}$ | 39 of 60 | 4 of 60 |
| Hierarchical versus GMM ($K = 30$) | 0.0015 | [0.0011, 0.0019] | 19/20 | $< 10^{-5}$ | 4 of 25 | 13 of 25 |
| Hierarchical ($K = 30$) versus HDBSCAN[a] | 0.014 | [0.012, 0.016] | 20/20 | $< 10^{-5}$ | — | — |

*Note:* Δ is the mean V-measure difference (first minus second method) with a bootstrap 95% confidence interval (CI). Wins is the number of subsamples in which the first method scored higher. $p$ is from the exact Wilcoxon signed-rank test, whose smallest attainable value at $n = 20$ is $1.9 \times 10^{-6}$. The last two columns count the model × method combinations in which the first or the second method is significantly ahead after Holm correction.
[a]DINOv3 + t-SNE only.

advantage over GMM at $K = 30$ is significant but negligible (0.0015). Within individual combinations, GMM is significantly ahead on PCA, kernel PCA and Isomap, Hierarchical on four t-SNE and UMAP combinations, and for DINOv3 with either manifold method neither is significantly better.

## 4.5 | Unsupervised Clustering: HDBSCAN Versus DBSCAN

For real-world scenarios where species count is unknown, unsupervised methods must balance three criteria: (1) discovering the true count of clusters, (2) minimising outliers requiring manual validation, and (3) achieving high clustering quality. Our results demonstrate HDBSCAN dramatically outperforms DBSCAN across all three dimensions (Table 5).

Using the best-performing configuration (DINOv3 + t-SNE), we evaluate both methods:

- Cluster count: HDBSCAN produces cluster counts approximating ground truth (30 species), discovering on average 33 clusters for birds and 37 for mammals. In contrast, DBSCAN over-splits dramatically, producing 256 clusters for birds and 247 for mammals, representing 8× over-clustering.
- Outlier ratio: HDBSCAN classifies only 0.9% (birds) and 1.4% (mammals) of images as outliers requiring manual review. DBSCAN rejects 28%–29% of samples across both taxonomic classes.
- Clustering quality: Despite operating without species count knowledge, HDBSCAN achieves V-measure of 0.948 (birds) and 0.939 (mammals), sacrificing only 1.3–1.5% points compared to the supervised Hierarchical $K = 30$ (V-measure = 0.961/0.954). DBSCAN achieves substantially lower V-measure of 0.668 for both classes.

This pattern persists across all model × dimensionality reduction combinations (Table A4), where DBSCAN consistently generates 150–260 clusters, while HDBSCAN maintains cluster counts within closer to ground truth.

Why DBSCAN fails: Automatic epsilon estimation, while theoretically sound, proves too sensitive to local density variations in the 2D embeddings, causing systematic over-fragmentation of species clusters.

## 4.6 | Consistent Cluster Count Prediction

From our results, we found that combining DINOv3, t-SNE and HDBSCAN we were able to use unsupervised methods to find clusters with high homogeneity, while getting close to the ground truth number of clusters. We evaluated this configuration's ability to predict the true cluster count by varying the number of species, $n \in \{5, 10, 15, 20, 25, 30\}$, for both birds and mammals. The selected species are randomized when $n < 30$, with 100 runs per $n$-value using identical pipeline settings (DINOv3 + t-SNE + HDBSCAN with min_cluster_size = 15, min_samples = 5), resulting in 1200 total runs (600 for mammals and 600 for birds).

As shown in Figure 6, the predicted cluster count closely follows the ground truth across all values of $n$. At $n = 30$, the method predicts 33.5 ± 2.1 clusters for birds and 37.5 ± 2.9 clusters for mammals, representing a slight but consistent overestimation.

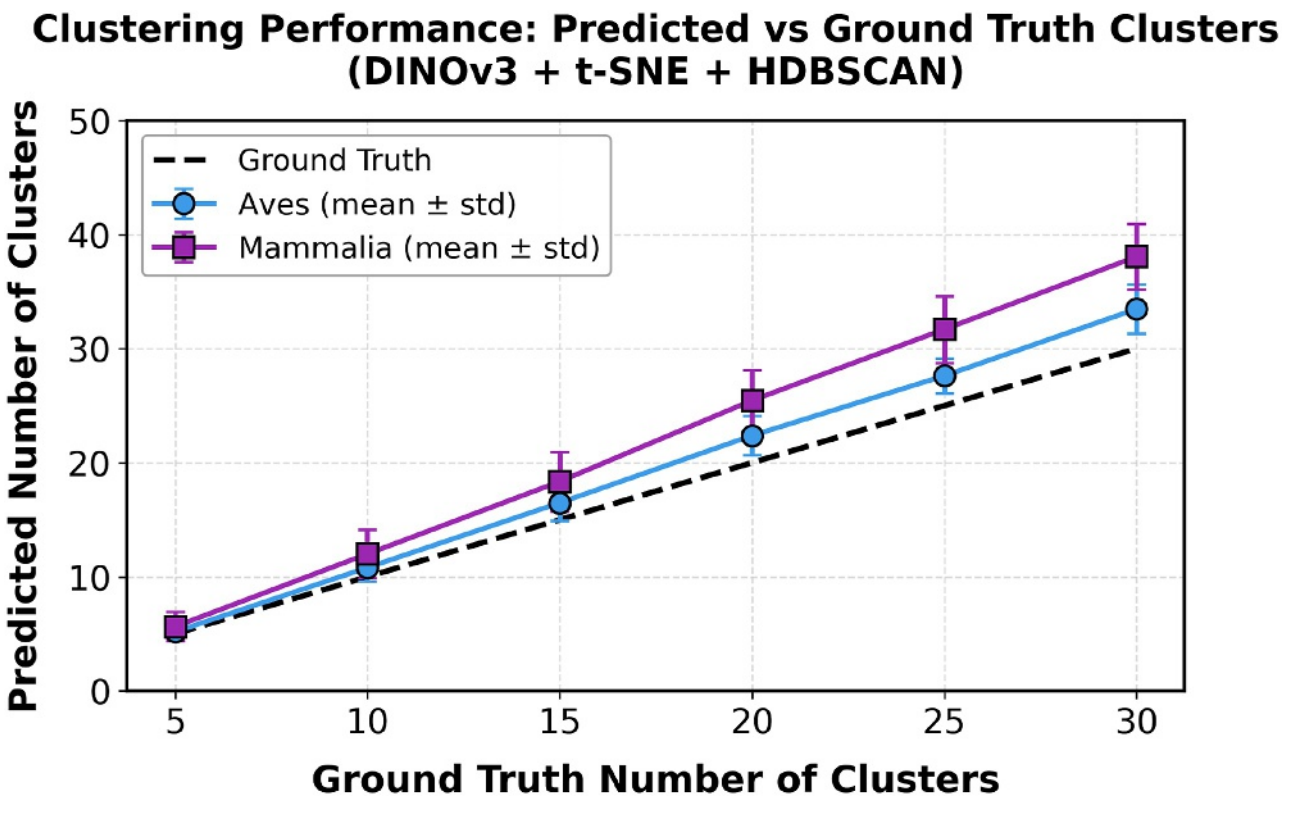

**FIGURE 6** | Mean predicted cluster count with standard deviation at different ground truth values $n \in \{5, 10, 15, 20, 25, 30\}$ for both mammals and birds, using DINOv3, t-SNE and HDBSCAN. The dashed line represents perfect prediction.

**TABLE 5** | Unsupervised clustering performance summary.

| Model | Dim. red. | Clusters B/M | Outliers B/M | V-measure B/M |
|---|---|---|---|---|
| HDBSCAN results | | | | |
| DINOv3 | t-SNE | 33/37 | 0.9%/1.4% | **0.948/0.939** |
| DINOv3 | UMAP | 47/57 | 4.2%/7.1% | 0.917/0.890 |
| DINOv2 | t-SNE | 34/45 | 2.3%/5.0% | 0.903/0.843 |
| DINOv2 | UMAP | 39/55 | 4.7%/8.3% | 0.889/0.802 |
| DBSCAN results | | | | |
| DINOv3 | t-SNE | 256/247 | 28%/29% | 0.668/0.668 |
| DINOv3 | UMAP | 204/189 | 31%/27% | 0.674/0.687 |
| DINOv2 | t-SNE | 221/218 | 28%/28% | 0.677/0.672 |
| DINOv2 | UMAP | 173/149 | 30%/26% | 0.680/0.685 |

*Note:* Averaged numbers. Ground truth: 30 clusters per class (B = Birds, M = Mammals). The bold marks the best configuration (highest V-measure).

#### 4.6.1 | Species-Level Clustering Behaviour

To understand clustering behaviour at the species level, we introduce two complementary metrics. The *Isolation Index* (II) measures the average purity experienced by each image and what fraction of its cluster-mates belong to the same species:

$$II_s = \frac{1}{N_s}\sum_{c\in C}\frac{n_{s,c}^2}{|c|} \tag{1}$$

The *Effective Cluster Count* (ECC) measures how many clusters a species effectively 'owns':

$$ECC_s = \sum_{c\in C}\frac{n_{s,c}}{|c|} \tag{2}$$

where $N_s$ is the total number of images of species $s$, $n_{s,c}$ is the count of species $s$ in cluster $c$, and $|c|$ is the cluster size. Together, these metrics classify species into distinct behavioural categories (Table 6).

The results reveal three distinct clustering behaviours. *Oversplit* species fragment into multiple pure clusters due to high intra-specific variation. Least Weasel exhibits seasonal pelage changes (brown summer coat vs. white winter coat), Red Junglefowl shows strong sexual dimorphism between colourful males and cryptic females, NZ Sea Lion displays pronounced size dimorphism between sexes, and Raccoon images split by imaging conditions (IR vs. RGB) and body postures. These splits, while inflating cluster counts, do not compromise identification accuracy, as each fragment remains species-pure and can be merged during post-processing.

*Merged* species pairs share clusters due to genuine visual similarity rather than model limitations: Wolf and Black-backed Jackal ($II \approx 0.63$) are both medium-sized canids with similar body proportions; Takahē and Australasian Swamphen ($II \approx 0.58$) are congeneric rails historically considered conspecific; Sun Bear shows the lowest ECC (0.85), indicating it is often absorbed into American Black Bear clusters rather than forming its own.

The majority of species (16/30 mammals, 14/30 birds) achieve *Ideal* clustering with $II \geq 0.95$ and $ECC$ close to 1.0, demonstrating that the pipeline successfully isolates most species.

Notably, some species exhibit both behaviours simultaneously: Red Junglefowl has high $ECC = 1.90$ (oversplit) but also low $II = 0.89$ (merging with Chicken), indicating it forms multiple clusters while also sharing some with its domesticated descendant.

## 4.7 | Intra-Specific Variation and Cluster Splitting

To assess whether additional clusters (> 30 per taxonomic class) could reveal ecologically meaningful intra-specific variation, we manually inspected clustering results from DINOv3 embeddings

**TABLE 6** | Species clustering behaviour at $n = 30$ using DINOv3 + t-SNE + HDBSCAN (100 runs).

| Species | II | ECC | Notes |
|---|---|---|---|
| *Oversplit*, multiple pure clusters | | | |
| Least Weasel (M) | 0.97 | **2.13** | Seasonal pelage variation |
| Raccoon (M) | 0.96 | **2.00** | IR/RGB, posture variation |
| Red Junglefowl (A) | 0.89 | **1.90** | Sexual dimorphism |
| NZ Sea Lion (M) | 0.95 | **1.84** | Age/sex dimorphism |
| Pipit (A) | 0.97 | **1.76** | Cryptic plumage variation |
| Petrel (A) | 0.91 | **1.71** | Flight versus ground poses |
| *Merged*, species pairs clustering together | | | |
| Takahē (A) | **0.58** | 0.83 | ↔ Swamphen |
| Swamphen (A) | **0.58** | 0.99 | ↔ Takahē |
| Wolf (M) | **0.63** | 1.78 | ↔ Jackal |
| Jackal (M) | **0.64** | 1.47 | ↔ Wolf |
| Black Bear (M) | **0.81** | 0.96 | ↔ Sun Bear |
| Sun Bear (M) | **0.82** | 0.85 | ↔ black Bear |
| *Ideal*, single pure cluster $ECC \approx 1.0$, $II \approx 1.0$ | | | |
| Gemsbok (M) | 1.00 | 1.00 | — |
| Porcupine (M) | 1.00 | 1.00 | — |
| Ostrich (A) | 1.00 | 1.00 | — |
| Kori Bustard (A) | 1.00 | 1.00 | — |
| + 26 Ideal species omitted (16M, 10A) | | | |

*Note:* Species are grouped by behaviour: *Oversplit* ($ECC \geq 1.5$, high II) fragment into multiple pure clusters; *Merged* ($II < 0.95$) share clusters with similar species; *Ideal* $II \geq 0.95$, $ECC \approx 1$ form single pure clusters. The bold marks the metric that places a species in its group: ECC ≥ 1.5 for oversplit species and II < 0.95 for merged species. Abbreviations: A, Aves; M, Mammalia.

reduced via t-SNE and clustered using both DBSCAN and supervised methods with $K \in \{90, 180\}$. While supervised methods force all embeddings into clusters, DBSCAN produced smaller, more homogeneous sub-clusters that consistently captured the same traits across independent runs.

Manual inspection methodology: We used the configuration that over-splits most, DINOv3 + t-SNE + DBSCAN, on the same 10 image subsets per taxonomic class as Table A4. DBSCAN produced 224–278 bird and 211–270 mammal clusters per run, splitting most species into several sub-clusters. One of the authors inspected every image of every sub-cluster for the species in Table 7 (643 sub-clusters over the 10 runs, with mean species purity between 0.981 and 1.000) and tagged a sub-cluster with a trait when all or nearly all of its images showed the same visible property, such as age class, sex, pelage, image modality or scene context. In total, 293 of the 643 sub-clusters (46%) showed such a trait, 165 (26%) a biological one (sex, age class, pelage or plumage, or a single individual) and 128 (20%) an image modality, scene or viewpoint. As one author did the inspection, no inter-rater agreement was measured, so this analysis is exploratory. The remaining sub-clusters may well hold further traits, behaviours or other patterns, and the traits reported here are those that were easy to distinguish by eye. Because each subset is a random sample, trait prevalence varied between runs, so Table 7 reports for each trait the number of runs in which a distinct sub-cluster formed.

Observed intra-specific patterns: While manually inspecting clusters, we identified several intra-species traits, such as juvenile, sex, fur and feather colouration and patterns, and clusters collecting images of body parts, where the cropping only included for example feet of the animal, rather than the whole animal. Additionally, we found clusters with different animal poses clustered together such as animals lying down. While looking for intra-species traits, we found clusters separating RGB, IR and white-light flash images across most species. At times, we also noticed clear patterns in clusters where the environmental background was dominant, such as clusters where snow was consistently present throughout.

We have selected a few species with observed traits, where we record the occurrence across 10 image runs and the results can be seen in Table 7. These traits and more, including more species, are available in the interactive web interface, provided as supplement to this paper (Section 6), where we aim to update and include more discoveries from the test data.

These results demonstrate that vision transformer embeddings can capture fine-grained visual differences beyond species. Age classes (juveniles), sexual dimorphism (males vs. females), phenotypic variation (fur/feather colour), imaging conditions (IR, flash and RGB), and environmental context (snow, posture) all produced separable clusters. We distinguish between two categories of detected patterns: biologically meaningful intra-specific variation and environmentally-driven clustering.

Intra-specific variation: Figures 7–9 present representative examples of detected traits, with images sampled from homogeneous clusters identified across independent runs. Figure 7 shows two examples: (a) male Greater Kudu in IR images, distinguished by their spiral horns; and (b) juvenile Kori Bustards showing characteristic immature plumage. Additional wolf-specific examples, including dark fur phenotypes and juvenile individuals, are presented in Figure 8. Quantitatively, the inspected sub-clusters were almost always species-pure (mean

**TABLE 7** | Intra-specific variation detected via over-clustering with DINOv3 + t-SNE + DBSCAN, the configuration from Table A4 that over-splits most, on the same 10 image subsets per taxonomic class.

| Species | Trait detected | Runs | II | ECC | Locations |
|---|---|---|---|---|---|
| Wolf | Juvenile individuals | 10/10 | 0.973 ± 0.008 | 6.7 ± 1.8 | 11.2 ± 7.3 |
| | Dark/black fur phenotype | 10/10 | | | |
| | IR (night) images | 10/10 | | | |
| | Snow background context | 10/10 | | | |
| Black-backed Jackal | Juvenile individuals | 5/10 | 0.984 ± 0.006 | 8.2 ± 2.6 | 12.5 ± 9.6 |
| | IR (night) images | 10/10 | | | |
| | White-light flash images | 10/10 | | | |
| Dromedary Camel | Leg-only (partial body) | 10/10 | 0.999 ± 0.002 | 11.2 ± 1.7 | 2.6 ± 1.9 |
| Greater Kudu | Males (IR images) | 10/10 | 0.995 ± 0.004 | 6.4 ± 2.0 | 5.1 ± 3.3 |
| | Males (RGB images) | 5/10 | | | |
| | Females + juveniles, lying posture | 5/10 | | | |
| Kori Bustard | Juvenile individuals | 5/10 | 1.000 ± 0.000 | 7.8 ± 2.7 | 6.6 ± 2.3 |
| Yellow-eyed Penguin | Juvenile individuals | 10/10 | 0.997 ± 0.005 | 8.6 ± 1.6 | 1.3 ± 0.5 |
| Chicken | Feather pattern/colour variants | 10/10 | 0.949 ± 0.018 | 13.0 ± 2.3 | — |
| | Males (sexual dimorphism) | 3/10 | | | |

*Note:* Fractions indicate runs where homogeneous trait clusters were identified. II and ECC quantify the same over-clustering run per species (mean ± sd over 10 runs, noise excluded) and show how consistently a species fragments. They are the Isolation Index and Effective Cluster Count of Section 4.6.1, where ECC counts the fragments the species splits into and II is the mean fraction of an image's cluster-mates that share its species. Locations is the mean ± sd number of camera locations contributing to one trait cluster. The Yellow-eyed Penguin images come from only three camera locations, hence its low value. For the Chicken, the camera location is unknown for 30% of the images.

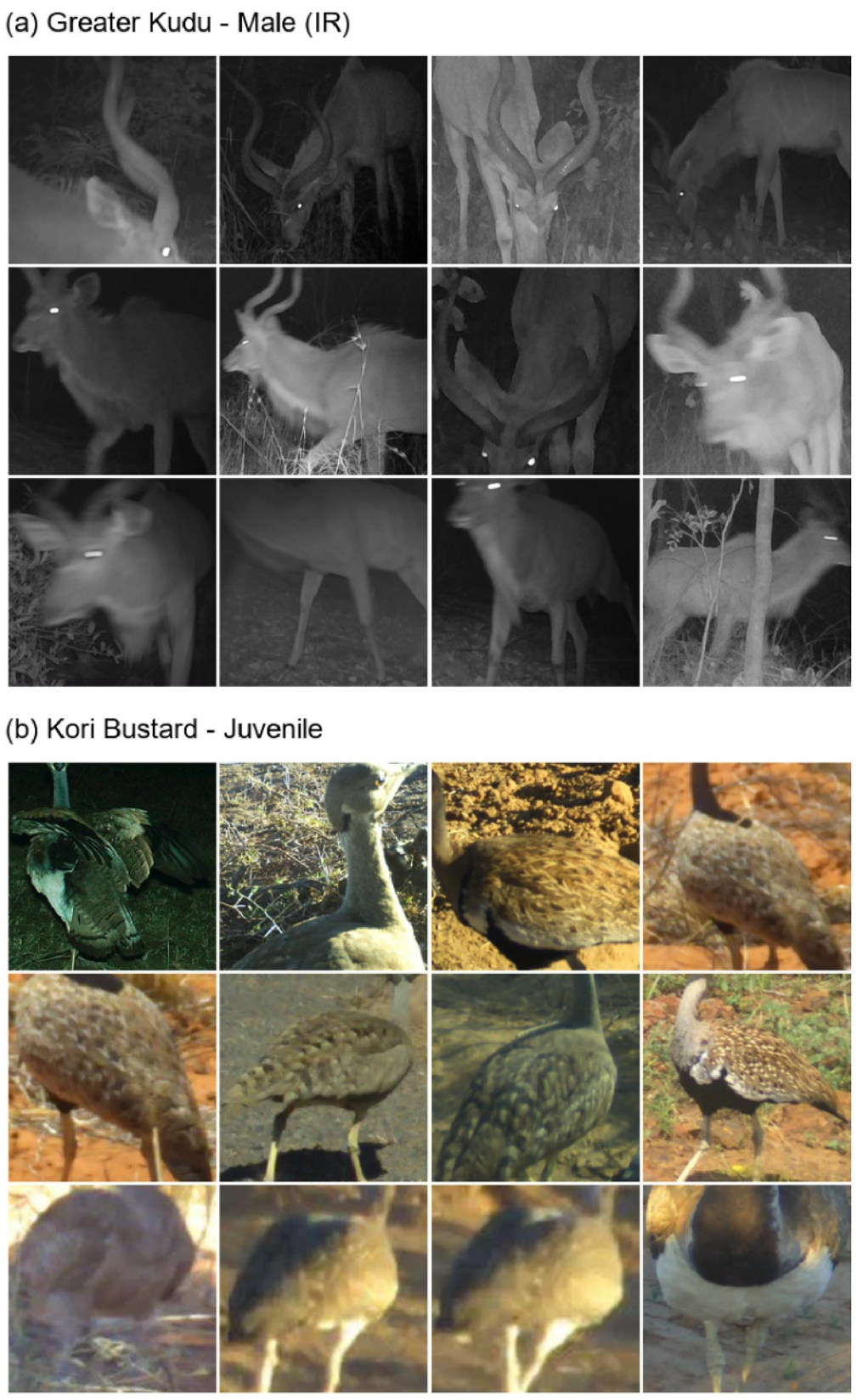

**FIGURE 7** | Examples of intra-specific variation detected via over-clustering (DINOv3 + t-SNE + DBSCAN). Each panel shows 12 images from a homogeneous cluster representing: (a) male Greater Kudu (IR); (b) juvenile Kori Bustards. Images were sampled from clusters identified across 10 independent runs with different random image subsets; occasional outliers may be present. Images are from the source datasets of each species: Greater Kudu [30, 27, 49] and Kori Bustard [38, 41, 42, 45, 46, 49].

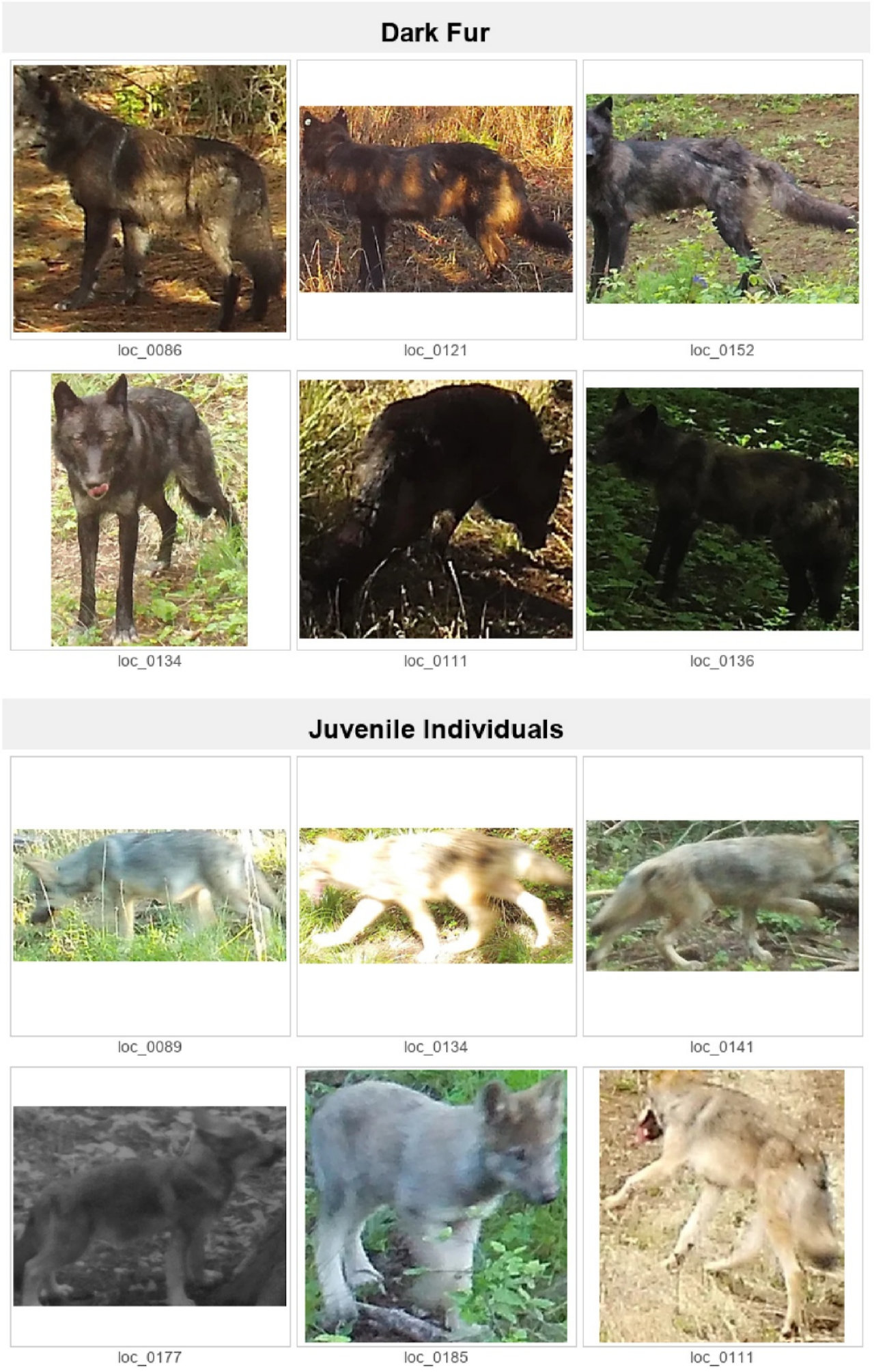

**FIGURE 8** | Location-agnostic clustering of biological traits. Samples from clusters of a single run (R1). Images from the Idaho Camera Traps dataset [32].

species purity 0.981 to 1.000 per species across 643 sub-clusters). Occasional outliers remain where the embedding similarity was driven by other visual features. The consistency of these clusters across multiple random subsets suggests that the detected patterns reflect genuine visual structure in the embedding space rather than sampling artifacts.

Environmental context clustering: Beyond biological traits, the embeddings also captured environmental and contextual features present in the images. Wolves photographed against snowy backgrounds were consistently grouped together across runs (see Figure 9, bottom panel). While such clusters do not represent intra-specific biological variation, they demonstrate the sensitivity of vision transformer embeddings to scene context and could be valuable for filtering images by environmental conditions or identifying seasonal patterns in camera trap data.

Location-agnostic clustering verification: A potential concern with camera trap datasets is that clustering might be driven by location-specific features, such as background vegetation, camera angle, or lighting conditions, rather than biologically meaningful traits. To address this concern, we traced each wolf image back to its original camera trap location. The Idaho camera trap dataset includes location identifiers in the original filenames (e.g., loc_0086, loc_0121), enabling systematic verification of location diversity within each trait cluster.

Table 8 shows the number of unique camera locations contributing to each wolf trait cluster across 10 independent runs. The dark fur cluster contains images from 24 different camera locations across the study area, with individual runs clustering from 4 to 13 locations. Similarly, infrared night images span 55 unique locations, juvenile clusters span 16 locations, and snowy background images originate from 31 locations.

Figures 8 and 9 present representative examples from each wolf trait cluster from run 1 (R1), where each of the six images originates from a different camera trap location. The visual coherence within each panel, despite the diverse geographic origins, demonstrates that DINOv3 embeddings capture genuine phenotypic and contextual patterns rather than location-specific artifacts.

To test whether this holds beyond wolves, we traced the trait clusters of Table 7 back to the camera locations of their images, using the original image paths of the source datasets and, for the WCS collection, its published location metadata Wildlife

Conservation Society [49]. The Locations column of Table 7 shows that trait clusters of the Black-backed Jackal and Wolf combine images from 11 to 13 camera locations on average, and those of the Greater Kudu and Kori Bustard from 5 to 7. Leg-only clusters of the Dromedary Camel draw on about 3 locations. The Yellow-eyed Penguin images come from only three camera locations in total, which explains the low value and small spread of its juvenile clusters.

## 4.8 | Performance on Unevenly Distributed Data

While our initial tests used 200 images per species, real-world camera-trap deployments typically exhibit long-tailed species distributions where common species vastly outnumber rare ones. We systematically evaluated whether this imbalance affects clustering quality.

We compared the original HDBSCAN (15, 5) configuration, combined with DINOv3 and t-SNE, across three distribution scenarios: even (200 species), mild imbalance (20–200 species), and extreme variation reflecting realistic deployment conditions (Table 9). The extreme scenario samples randomly between 20 and the maximum available images per species, resulting in ranges of 48–4145 for Aves and 29–6431 for Mammalia. The lower limit of 20 images keeps every species above the min_cluster_size of 15 used by HDBSCAN (15, 5), and the upper limit of 200 images matches the even distribution used throughout the paper, whereas the extreme scenario removes the upper limit.

The mild imbalance (20–200 images per species) shows *no degradation*, in fact, cluster estimation slightly improves. However, extreme imbalance causes significant over-clustering: HDBSCAN produced an average of 415 clusters for Aves and 304 for Mammalia instead of the target 30, representing 10–14× over-clustering. While homogeneity remains high under extreme imbalance (see Appendix Table A6), this comes at the cost of substantially more outliers and fragmented clusters requiring manual review.

To locate where performance starts to degrade, we repeated the uneven sampling with upper limits from 20 to 3200 images and with all available images, which reproduces the extreme scenario (Figure 10). HDBSCAN (15, 5) performs equally well up to an upper limit of 400 images (V-measure 0.94–0.95), which covers the whole mild range, and degrades from 800 images onwards (0.77 with all images) as the largest species are split into several clusters. The larger configurations tested next show the opposite behaviour. They remain stable on extreme data (0.93–0.94) but miss species with fewer images than their `min_cluster_size`, and most images of such species are labelled as outliers for expert review (Table 10). The configuration should therefore be chosen by the size of the largest species, with HDBSCAN (15, 5) suited to datasets in which no species has more than about 400 images.

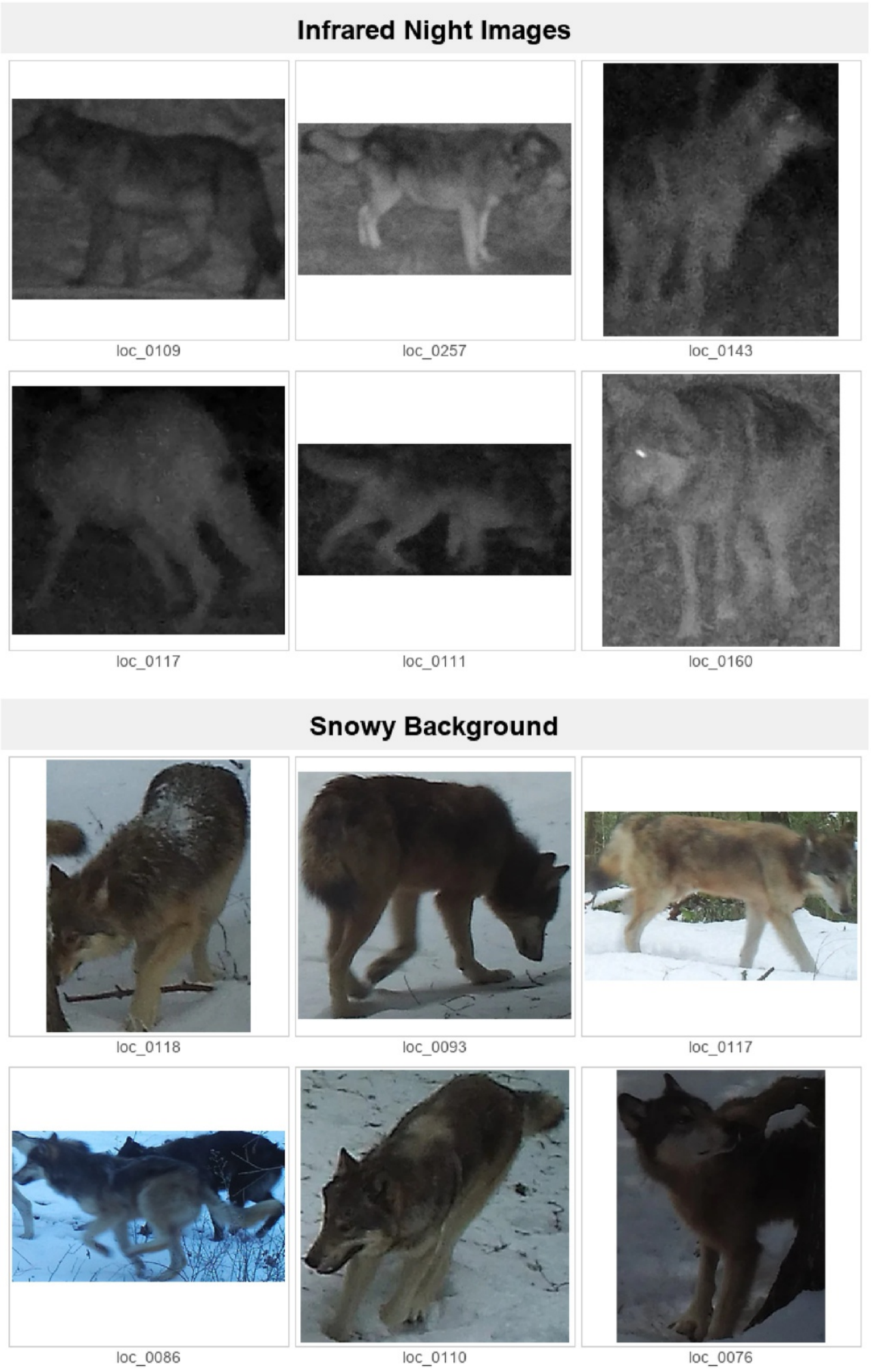

**FIGURE 9** | Location-agnostic clustering of IR images and images with snowy backgrounds with wolves. Samples from clusters of a single run (R1). Images from the Idaho Camera Traps dataset [32].

### 4.8.1 | *Parameter Optimization for Extreme Imbalance*

To address the over-clustering issue, we tested configurations with larger `min_cluster_size` values for HDBSCAN and larger `min_samples`/epsilon multipliers for DBSCAN (see Appendix D for comprehensive results across 20 configurations). Table 11 summarises the two best-performing

**TABLE 8** | Location distribution for wolf trait clusters across 10 independent runs.

| Trait | R1 | R2 | R3 | R4 | R5 | R6 | R7 | R8 | R9 | R10 |
|---|---|---|---|---|---|---|---|---|---|---|
| Dark Fur | 6 | 13 | 7 | 5 | 4 | 11 | 9 | 11 | 8 | 10 |
| Juvenile | 8 | 4 | 6 | 2 | 2 | 3 | 5 | 5 | 5 | 5 |
| Night images (IR) | 15 | 16 | 16 | 19 | 23 | 19 | 22 | 23 | 30 | 29 |
| Snowy background | 9 | 9 | 7 | 8 | 8 | 13 | 10 | 17 | 12 | 13 |

*Note:* Each cell shows the number of unique camera locations contributing to that trait cluster.

TABLE 9 | Effect of sample distribution on clustering with 30 species (DINOv3 + t-SNE).

| Method | Aves | | Mammalia | |
|---|---|---|---|---|
| | V-measure | Clusters | V-measure | Clusters |
| HDBSCAN (15, 5) | | | | |
| Even | 0.948 ± 0.006 | 33 ± 3 | 0.939 ± 0.006 | 37 ± 3 |
| Uneven | 0.954 ± 0.006 | 31 ± 1 | 0.950 ± 0.009 | 33 ± 2 |
| Extreme | 0.776 ± 0.029 | **415 ± 92** | 0.792 ± 0.022 | **304 ± 51** |

*Note:* Even: 200 img/sp. Uneven: 20–200 img/sp. Extreme: random 20–max img/sp. The bold marks problematic values (cluster counts far above the 30 true species).

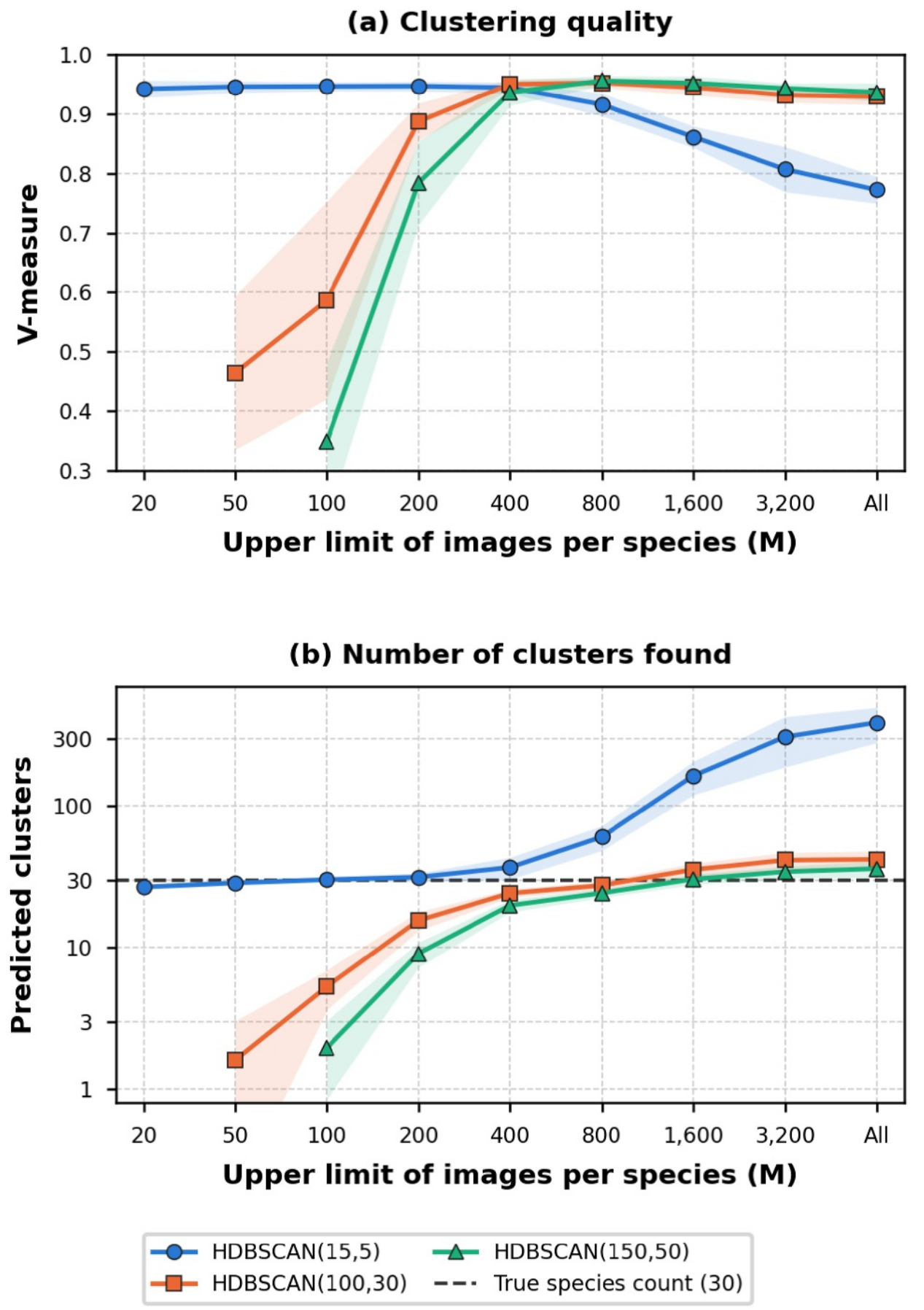

**FIGURE 10** | Transition from mild to extreme imbalance (DINOv3 + t-SNE, mean over 20 runs, 10 per taxonomic class, shaded band ±1 SD). Each species receives a random number of images between 20 and the upper limit $M$. $M = 20$ gives every species exactly 20 images, $M = 200$ is the mild scenario and $M =$ All reproduces the extreme scenario. (a) V-measure. (b) Number of predicted clusters, on a log scale. Points are omitted where a configuration found no clusters.

configurations for each clustering method, including results on combined bird and mammal data.

When balancing V-measure, cluster count accuracy, and outlier rates, HDBSCAN (150, 50) emerged as the most effective configuration for extreme uneven distributions, achieving V = 0.948 with t-SNE and V = 0.942 with UMAP on the combined 60-species dataset. The DBSCAN (2.0, 30) configuration also performs well with the lowest outlier rate (0.5% for t-SNE),

TABLE 10 | Fate of species with fewer images than min_cluster_size threshold. Shows percentage classified as outliers versus merged into other clusters.

| Method | Dataset | HDBSCAN (100, 30) | | HDBSCAN (150, 50) | |
|---|---|---|---|---|---|
| | | Outlier | Merged | Outlier | Merged |
| t-SNE | Aves | 71% | 29% | 70% | 30% |
| t-SNE | Mammalia | 69% | 31% | 74% | 26% |
| t-SNE | Combined | 59% | 41% | 60% | 40% |
| UMAP | Aves | 22% | 78% | 37% | 63% |
| UMAP | Mammalia | 60% | 40% | 48% | 52% |
| UMAP | Combined | 37% | 63% | 41% | 59% |

though it occasionally produces fewer clusters than the true species count.

### 4.8.2 | *Scaling Behaviour*

To further evaluate the scalability of the configurations optimised for extreme imbalance, we tested them across varying numbers of species: $n = 5, 10, \ldots, 30$ for single-class datasets and $n = 5, 10, \ldots, 60$ for combined datasets. Each test used 10 random species samples with 20–MAX images per species.

Both HDBSCAN configurations showed minimal scaling degradation in V-measure, which stayed above 0.92 from 10 species upwards, although their predicted cluster counts rose above the true species count as species were added, to 74 for HDBSCAN (150, 50) and 89 for HDBSCAN (100, 30) at 60 species (Figure 11). The two DBSCAN configurations exhibited scaling limitations. DBSCAN (1.5, 20) achieved comparable homogeneity to the HDBSCAN configurations but consistently over-clustered, reducing V-measure. DBSCAN (2.0, 30) showed promising results with UMAP but tended to under-cluster with t-SNE, lowering homogeneity.

### 4.8.3 | *Generalisation Across Distribution Types*

To verify that optimised configurations generalise beyond their tuning scenario, we applied all methods to both even and extreme distributions (Table 12).

DBSCAN configurations optimised for extreme imbalance fail on evenly distributed data, with 72%–78% of images classified as outliers. This occurs because the larger epsilon multipliers tuned for the spread-out extreme distributions over-smooth the denser, more uniform clusters in even data, causing most points to fall outside cluster boundaries.

In contrast, both optimised HDBSCAN configurations demonstrate robust performance across distribution types. HDBSCAN (100, 30) achieves V-measure of 0.947–0.951 on even data while maintaining 0.918–0.936 on extreme distributions. HDBSCAN (150, 50) shows remarkable consistency, actually performing *better* on extreme Aves data (V = 0.946) than on even data (V = 0.943), while keeping outlier rates below 5%.

**TABLE 11** | Best configurations for extreme uneven distributions (20–MAX images per species). Results averaged over 10 different image distributions. Full results in Appendix Table A6.

| Config | t-SNE | | | UMAP | | |
|---|---|---|---|---|---|---|
| | V-m | Cl. | O% | V-m | Cl. | O% |
| Aves (30 species) | | | | | | |
| HDBSCAN (100, 30) | 0.936 | 43 | 4.7 | 0.922 | 45 | 3.0 |
| HDBSCAN (150, 50) | **0.946** | 36 | 4.7 | 0.933 | 36 | 2.3 |
| DBSCAN (1.5, 20) | 0.909 | 75 | 2.3 | 0.910 | 80 | 2.9 |
| DBSCAN (2.0, 30) | 0.911 | 28 | 0.3 | **0.940** | 45 | 1.6 |
| Mammalia (30 species) | | | | | | |
| HDBSCAN (100, 30) | 0.918 | 44 | 3.9 | 0.920 | 45 | 2.1 |
| HDBSCAN (150, 50) | **0.930** | 39 | 3.7 | **0.926** | 40 | 1.9 |
| DBSCAN (1.5, 20) | 0.887 | 68 | 2.1 | 0.877 | 85 | 3.9 |
| DBSCAN (2.0, 30) | 0.899 | 32 | 0.2 | 0.914 | 50 | 1.1 |
| Combined (60 species) | | | | | | |
| HDBSCAN (100, 30) | 0.936 | 91 | 4.2 | 0.933 | 89 | 3.1 |
| HDBSCAN (150, 50) | **0.948** | 72 | 4.1 | **0.942** | 74 | 2.3 |
| DBSCAN (1.5, 20) | 0.900 | 167 | 2.8 | 0.910 | 155 | 4.3 |
| DBSCAN (2.0, 30) | 0.930 | 68 | 0.5 | 0.939 | 95 | 1.6 |

*Note:* The bold marks the highest V-measure per dataset and dimensionality reduction method.
Abbreviations: Cl. = Clusters found (true: 30/30/60), O% = Outlier ratio, V-m = V-measure.

This asymmetry has important practical implications: HDBSCAN configurations can be safely applied without prior knowledge of data distribution, whereas DBSCAN requires distribution-specific tuning and should be avoided when distribution characteristics are unknown.

In HDBSCAN, min_cluster_size specifies the minimum number of images required to form a cluster. With HDBSCAN (150, 50), species having fewer than 150 images cannot form their own cluster and will either be merged into a visually similar larger cluster or become outliers requiring manual review.

#### 4.8.4 | *Fate of Under-Represented Species*

A critical question for practical deployment is what happens to species with fewer images than the min_cluster_size threshold. We tracked outcomes for all such species across our scaling experiments (Table 10).

The choice of dimensionality reduction significantly affects how under-represented species are handled. With t-SNE, approximately 60%–74% of under-threshold species become outliers, while 26%–41% merge into larger clusters. UMAP shows the opposite pattern: only 22%–48% become outliers, with 52%–78% merging into visually similar clusters.

This difference has practical implications for workflow design. If the goal is to flag rare species for expert review, t-SNE's higher outlier rate is advantageous, rare species are more likely to appear in the outlier pool requiring manual inspection. If the goal is to minimise data requiring manual attention (accepting that rare species may be grouped with visually similar species), UMAP's tendency to merge is preferable. In both cases, the

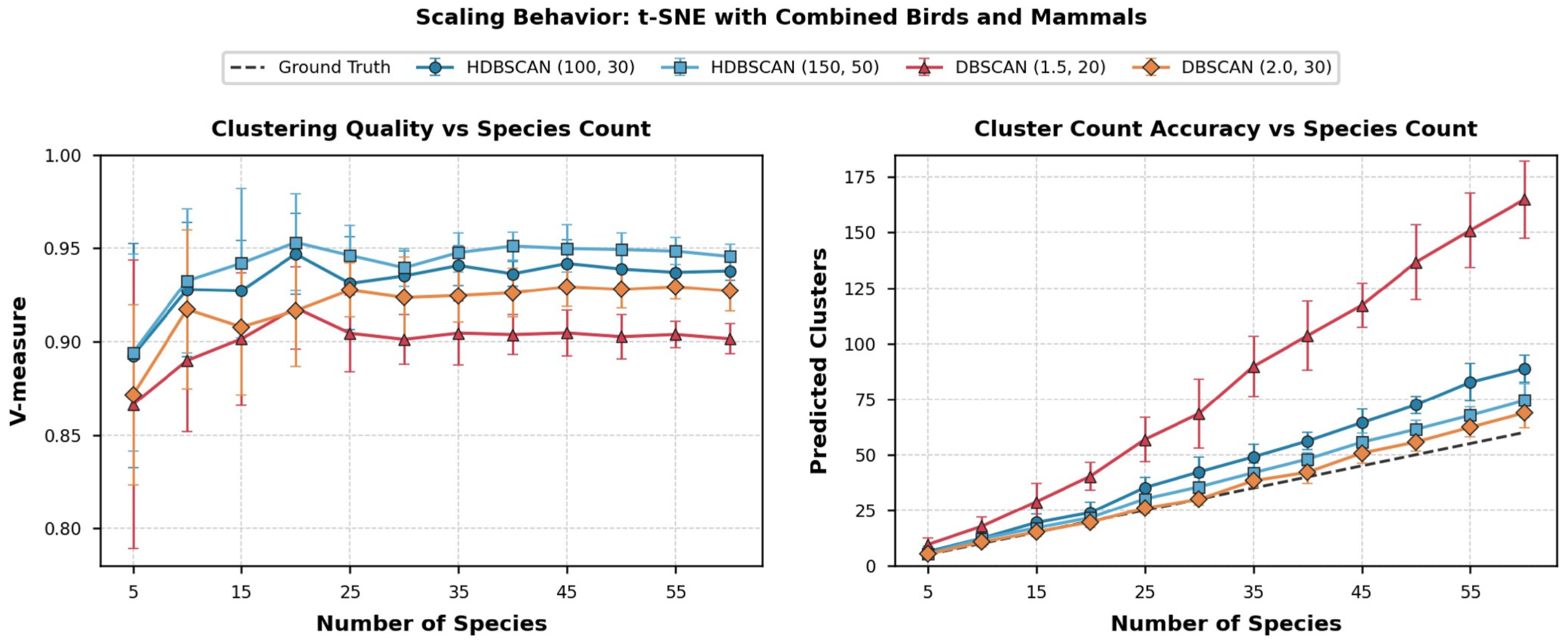

**FIGURE 11** | Scaling behaviour of unsupervised clustering using t-SNE on combined bird and mammal datasets (DINOv3 embeddings). The left panel shows that V-measure stays above 0.92 for both HDBSCAN configurations from 10 species upwards. In the right panel, DBSCAN (2.0, 30) and HDBSCAN (150, 50) track the true species count (dashed line) most closely, while DBSCAN (1.5, 20) increasingly over-splits as species are added. Error bars show standard deviation across 10 experimental runs. See Figure A1 in Appendix D for complete results across the selected configurations.

**TABLE 12** | Clustering performance across distribution types (DINOv3 + t-SNE, 30 species).

| Method | Aves | | | Mammalia | | |
|---|---|---|---|---|---|---|
| | **V-m** | **Cl.** | **O%** | **V-m** | **Cl.** | **O%** |
| HDBSCAN (15, 5)—Original | | | | | | |
| Even | 0.948 ± 0.006 | 33 ± 3 | 0.9 | 0.939 ± 0.006 | 37 ± 3 | 1.4 |
| Extreme | 0.776 ± 0.029 | **415 ± 92** | 14.2 | 0.792 ± 0.022 | **304 ± 51** | 12.2 |
| HDBSCAN (100, 30)—Optimised | | | | | | |
| Even | 0.951 ± 0.007 | 29 ± 2 | 1.1 | 0.947 ± 0.004 | 29 ± 2 | 1.9 |
| Extreme | 0.936 ± 0.012 | 43 ± 4 | 4.7 | 0.918 ± 0.015 | 44 ± 3 | 3.9 |
| HDBSCAN (150, 50)—Optimised | | | | | | |
| Even | 0.943 ± 0.009 | 27 ± 2 | 1.2 | 0.926 ± 0.019 | 26 ± 3 | 1.6 |
| Extreme | 0.946 ± 0.009 | 36 ± 3 | 4.7 | 0.930 ± 0.015 | 39 ± 3 | 3.7 |
| DBSCAN (1.5, 20)—Optimised for Extreme | | | | | | |
| Even | 0.324 | 41 | **77.6** | 0.324 | 41 | **77.6** |
| Extreme | 0.909 | 75 | 2.3 | 0.887 | 68 | 2.1 |
| DBSCAN (2.0, 30)—Optimised for Extreme | | | | | | |
| Even | 0.383 | 34 | **72.5** | 0.383 | 34 | **72.5** |
| Extreme | 0.911 | 28 | 0.3 | 0.899 | 32 | 0.2 |

*Note:* Bold indicates problematic values. Even: 200 img/sp. Extreme: 20–max img/sp.
Abbreviations: Cl. = Clusters found (target: 30), O% = Outlier ratio, V-m = V-measure.

optimised configurations maintain high overall V-measure (> 0.92) while providing interpretable behaviour for edge cases.

#### 4.8.5 | *Practical Recommendations*

Our results demonstrate that DINOv3 combined with t-SNE and HDBSCAN provides robust species-level clustering across distribution types, with UMAP offering a viable alternative with lower outlier rates. The ideal deployment produces high-homogeneity clusters approximating the true species count, with outliers capturing rare species and ambiguous images for targeted manual inspection.

Parameter selection should be guided by expected data characteristics:

- Large datasets (> 1000 images, > 150 images per expected species): HDBSCAN (150, 50) provides optimal balance of cluster purity, accurate cluster count estimation, and manageable outlier rates (3%–5%).
- Moderate datasets or mixed representation: HDBSCAN (100, 30) offers a middle ground, maintaining strong performance on both even and uneven distributions.
- Smaller datasets or expected rare species (< 100 images for some species): The original HDBSCAN (15, 5) configuration prevents rare species from being systematically excluded from clustering, though at the cost of potential over-clustering with extreme imbalance.
- Prioritising rare species detection: Combine t-SNE with higher min_cluster_size to maximise the proportion of under-represented species classified as outliers for expert review.

DBSCAN configurations, while effective when tuned for specific distributions, should be avoided when distribution characteristics are unknown or variable, as they fail to generalise across distribution types. Measured runtimes and memory requirements for the full pipeline on a single consumer GPU and CPU are reported in Appendix F.

### 4.9 | Effect of Dimensionality

We conducted testing comparing clustering directly on raw high-dimensional embeddings (768D–1536D) against 2D t-SNE-reduced spaces (Table 13). Dimensionality reduction consistently improved clustering performance across all models and methods.

For HDBSCAN, t-SNE reduction improved the mean V-measure from 0.498 (raw) to 0.783, an absolute gain of 0.285. Vision-language models benefited most dramatically: CLIP improved from 0.212 to 0.697, and SigLIP from 0.196 to 0.671. DINO models showed smaller but consistent improvements (DINOv3: 0.819–0.943).

For Hierarchical clustering at $K = 30$, t-SNE reduction improved the mean V-measure from 0.432 to 0.784, an absolute gain of 0.352. However, this comparison is affected by algorithm failures, as Hierarchical clustering failed entirely on raw high-dimensional DINOv2 and BioCLIP 2 embeddings due to covariance matrix singularity but worked reliably on all t-SNE-reduced embeddings.

DBSCAN exhibited opposite behaviour, performing better on raw embeddings (DINOv3: mean of 38 clusters with V-measure of 0.693) while severely over-clustering with t-SNE (mean of 251 clusters with V-measure of 0.668), suggesting the automatic epsilon estimation is more stable in high-dimensional spaces where density variations are less pronounced. Intermediate

**TABLE 13** | Effect of dimensionality on clustering performance.

| Model | Dim. | HDBSCAN | | Hierarchical ($K = 30$) | |
|---|---|---|---|---|---|
| | | Raw | t-SNE | Raw | t-SNE |
| DINOv3 | 1280D | 0.819 | 0.943 | 0.920 | 0.958 |
| DINOv2 | 1536D | 0.745 | 0.873 | 0.000[a] | 0.879 |
| BioCLIP 2 | 768D | 0.519 | 0.730 | 0.000[a] | 0.737 |
| CLIP | 768D | 0.212 | 0.697 | 0.629 | 0.688 |
| SigLIP | 768D | 0.196 | 0.671 | 0.611 | 0.661 |
| Mean | — | 0.498 | 0.783 | 0.432[b] | 0.784 |

*Note:* Comparison of clustering directly on raw high-dimensional embeddings versus t-SNE-reduced embeddings (2D). V-measure reported as mean across 10 Aves and 10 Mammalia subsamples. Values of 0.000 indicate algorithm failure due to covariance matrix singularity.
[a]Algorithm failed due to covariance matrix singularity.
[b]Mean excludes failed runs (DINOv2, BioCLIP 2).

target dimensions (2D to 256D) are evaluated separately in Appendix E.

## 5 | Discussion

### 5.1 | Reduced Annotation Burden

The benchmark allows an estimate of how much manual labelling effort zero-shot clustering saves. A controlled study of annotation time was beyond the scope of this work, so we count labelling actions instead, assuming an annotation tool that assigns one label to all images of a cluster at once. With DINOv3, t-SNE and HDBSCAN, a bird and a mammal subset of 6000 images each form on average 33 and 37 clusters, 70 in total, and leave 0.9% and 1.4% of their images as outliers, on average 137 images or 1.14% (Table 5). Labelling each cluster once and each outlier individually therefore takes an average of 207 labelling actions instead of 12,000, a reduction of 98.3%. If each cluster receives the label of its most common species, 94.6% of the clustered images in the 20 benchmark subsets are labelled correctly. This is slightly below the 96.6% of individual experts and the 97.9% of aggregated volunteer labels in Snapshot Serengeti Swanson et al. [65], but within the 83.4%–97.9% reported across three citizen science projects Egna et al. [66]. Experts also disagreed on 39% of images of two similar lynx species Gooliaff and Hodges [67], which shows that morphologically similar species are hard for human annotators as well, and these are also the cases where zero-shot clustering most often fails (Section 4.6.1). The time saved is not measured here. Checking a cluster may take longer than checking a single image, although several images of the same species can also make an identification easier than one poorly visible image.

### 5.2 | Reduction to Higher Dimensions

Throughout our main benchmark, we either kept the raw embedding dimensions (768D–1536D depending on model architecture) or reduced them to 2D, which suits visualisation and manual cluster inspection. To test whether intermediate dimensionalities cluster better, we reduced DINOv3 embeddings to 2D–256D with UMAP, and to 2D and 3D with t-SNE, the output dimensions supported by its Barnes-Hut approximation (Appendix E).

With HDBSCAN, UMAP improves from a V-measure of 0.90 at 2D to 0.94 at 32D and then plateaus, mainly because its predicted cluster count approaches the true number of species. t-SNE reaches 0.94 at 2D and 0.93 at 3D, so from 8D upwards UMAP matches or slightly exceeds 2D t-SNE.

Across the main benchmark, t-SNE is slightly ahead of UMAP in 2D (Table 4), although this may partly reflect its projection rather than better separation of the species (Section 3.3.3). The two reducers also handle rare species differently. With t-SNE, most species that are too small to form their own cluster end up as outliers for expert review (60%–74%), whereas UMAP merges most of them into clusters of visually similar species (52%–78%, Table 10). t-SNE is therefore the better choice for clustering and visualisation in 2D, particularly when rare species must not be missed. When more of the information in the embeddings should be kept for downstream tasks, by retaining more dimensions, UMAP is the recommended reducer.

### 5.3 | Configuration Importance

Our results demonstrate that clustering algorithm performance is parameter-dependent, and optimal configurations vary with data characteristics. While HDBSCAN Campello et al. [24] consistently outperformed DBSCAN across our benchmark, this finding is contingent on parameter selection and on the datasets evaluated, since benchmarks on a limited set of datasets can bias conclusions about clustering methods Gagolewski [68].

Our systematic evaluation of parameter configurations (Table A6) revealed that the original HDBSCAN (15, 5) configuration, while optimal for evenly distributed data (V-measure = 0.943), fails dramatically on extreme uneven distributions, producing 300–400 clusters instead of the target 30. Conversely, configurations optimised for imbalanced data, HDBSCAN (100, 30) and HDBSCAN (150, 50), generalise well to both distribution types, maintaining V-measure $> 0.92$ across scenarios.

DBSCAN exhibited more severe sensitivity: configurations tuned for extreme imbalance classified 72%–78% of images as outliers when applied to evenly distributed data (Table 12). This asymmetry has critical practical implications: HDBSCAN parameters can be selected conservatively for unknown distributions, whereas DBSCAN requires distribution-specific tuning and should be avoided when data characteristics are uncertain. This is consistent with the stated design goal of the HDBSCAN implementation, to find clusters of varying densities and be more robust to parameter selection McInnes et al. [69].

We recommend practitioners: (1) use HDBSCAN over DBSCAN for robustness to unknown distributions; (2) set min_cluster_size based on the minimum expected images per species, HDBSCAN (150, 50) for large datasets, HDBSCAN (100, 30) for moderate ones; and (3) validate parameter choices on subsets when possible before full deployment.

### 5.4 | Clustering Failure Analysis

Our quantitative analysis of species-level clustering behaviour (Table 6) reveals systematic failure patterns that reflect genuine visual similarity rather than model limitations. Species pairs with low Isolation Index (II $< 0.70$) consistently share taxonomic or morphological similarity: Wolf and Black-backed Jackal (II $\approx 0.63$) are both medium-sized canids with similar body proportions; Takahē and Australasian Swamphen (II $\approx 0.58$) are congeneric rails historically considered conspecific; American Black Bear and Sun Bear (II $\approx 0.81$) share ursid body morphology despite geographic separation. Similar species pairs also challenge human annotators. Experts classifying camera-trap images of bobcat and Canada lynx reached moderate agreement with each other Gooliaff and Hodges [67]. A reply study showed that with training and several images per camera-trap sequence, even non-experts could separate the two species consistently Thornton et al. [70]. In Snapshot Serengeti, residual volunteer errors likewise include a congeneric gazelle pair Swanson et al. [46].

These merging failures are exacerbated by imaging conditions. IR (infrared) images, which lack colour information and often capture animals at distance, showed particularly high inter-species confusion. Night images of canids, for instance, reduced distinguishing features to silhouette and relative size, features that overlap substantially between wolves and jackals, a difficulty also reported for wolves, foxes and dogs in nocturnal camera-trap images Mulero-Pázmány et al. [71]. Similarly, black-feathered bird species showed elevated confusion rates in low-light conditions where plumage details were obscured.

DINOv3 embeddings generally achieved better species separation than other models, but imperfect separation persists for visually similar taxa. Notably, the embedding space retains meaningful structure even within merged clusters. We observed that images of different species within a shared cluster tend to occupy distinct sub-regions, with species boundaries often visible along embedding coordinate gradients. This suggests a practical workflow enhancement: rather than presenting merged cluster images in random order for manual review, sorting by embedding coordinates (x/y position in 2D space) can facilitate faster species separation by grouping visually similar images together. While manual validation remains necessary for merged clusters, coordinate-based sorting reduces workload by exploiting the partial separation preserved in the embedding geometry.

Oversplit species (ECC $> 1.5$) present a different challenge. Species like Least Weasel (ECC $= 2.13$) and Raccoon (ECC $= 2.00$) fragment into multiple pure clusters, likely reflecting high intra-specific variation in pelage, posture, or imaging conditions. These splits, while inflating cluster counts, do not compromise identification accuracy, each fragment remains species-pure and can be merged during post-processing.

### 5.5 | Spatial Organisation Patterns

For the two DINO models and BioCLIP 2, on evenly distributed species data, using t-SNE, we noticed from the 2D plots that clusters with high heterogeneity were often closer to the center of the plot, while more homogeneous clusters were located further away from the center and clustered more densely. We treat this as a speculative observation. Cluster sizes and distances between clusters in a t-SNE map are shaped by the algorithm and should not be read as properties of the data Wattenberg et al. [57], and a similar centre-versus-periphery layout, driven by how frequent each type of sample is, has been reported for t-SNE maps of solar spectra Verma et al. [72]. We therefore do not recommend distance from the plot origin as an indicator of cluster homogeneity. Confirming whether the pattern reflects the data would require validation on held-out data across random seeds and embedding models.

### 5.6 | Broader Applicability and Future Directions

While this study focused on birds and mammals, we hypothesise that zero-shot clustering generalises to other taxonomic groups. Camera trap and underwater monitoring systems increasingly target fishes, insects, reptiles and amphibians, all of which present similar challenges of high image volume and limited annotation capacity. Detection models similar to MegaDetector already exist for fish (Community Fish Detector WildHackers [6]) and insects (FlatBug Svenning et al. [7]), enabling the same crop-then-cluster pipeline evaluated here. The self-supervised nature of DINOv3 embeddings, trained on general visual features rather than taxon-specific datasets, suggests broad applicability, though performance may vary with morphological distinctiveness and imaging conditions specific to each domain, as reported for insect classification under lab-to-field domain shift Rayeed et al. [73] and for DINO representations transferred to underwater imagery Chen et al. [74].

A key direction for future work is systematically separating biological traits from environmental factors within the embedding space. The embedding structure this requires has been reported for BioCLIP 2, where intra-species variation is preserved in subspaces orthogonal to inter-species distinctions Gu et al. [13]. Our results (Table 7) show that DINOv3 captures both types of variation, which is valuable for ecologists, that may want to filter by imaging conditions or analyse population demographics depending on their research goals.

Additionally, hierarchical clustering approaches that first separate species, then systematically subdivide clusters to reveal intra-specific structure, could provide ecologists with multi-resolution views of their data. Such workflows would enable both rapid species-level sorting and detailed population demographic analysis from the same embedding space, maximising the utility of foundation model representations for biodiversity monitoring.

## 6 | Open-Source Contributions and Data Availability

Together with the publication of this paper, we release the full dataset of 139,111 annotated cropped images, together with an interactive demo to visualize some combinations of the different methods used, to visualize the 2D space the embedding points and the corresponding clusters and images. We also release benchmarking code in GitHub for anyone to reproduce and test the experiments we conducted, while giving the possibility to run the same tests with other ViT models and other types of data. Finally, we give access to all the collected logs and configurations tested used in this paper. All the contributions can be found in https://hugomarkoff.github.io/animal_visual_transformer/.

## 7 | Conclusion

This study addressed a fundamental challenge in biodiversity monitoring: whether Vision Transformer foundation models can reduce manual annotation burden by automatically clustering unlabelled camera trap images into species-level groups. Through systematic evaluation of 27,600 experimental configurations across 60 species and 139,111 validated images, we demonstrate that zero-shot clustering provides a practical pathway to accelerate ecological image analysis workflows.

Our benchmarking reveals that model selection has the greatest impact on clustering performance. For practical deployment where species counts are unknown, unsupervised HDBSCAN clustering of DINOv3 embeddings reaches a V-measure of 0.943 while discovering cluster counts within 18% of ground truth and rejecting only 1.14% of images as outliers requiring expert review. In practice, an annotator labels about 70 clusters and reviews 137 outliers instead of 12,000 single images, and the resulting labels are nearly as accurate as those of human experts. The remaining errors fall mostly on morphologically similar species, which also challenge human annotators (Section 5.1). When the true species count is supplied, supervised hierarchical clustering of DINOv3 embeddings reaches an upper bound of 0.958, substantially outperforming biology-specific BioCLIP 2 at 0.737 and general-purpose vision-language models such as CLIP and SigLIP. This result suggests that self-supervised learning on diverse visual data produces more transferable representations than the domain-specific training for clustering tasks. Dimensionality reduction proves essential, with t-SNE and UMAP improving clustering performance by 26–38% points compared to other tested methods. In 2D, t-SNE clusters slightly better than UMAP, whereas for workflows that retain more dimensions for downstream analysis UMAP is the better choice.

Beyond quantitative metrics, our ecological interpretation reveals that clustering failures often encode meaningful biological structure. Species pairs that cluster together, such as Wolf and Black-backed Jackal or Takahē and Australasian Swamphen, share genuine morphological or taxonomic similarity rather than representing model limitations. Conversely, over-split species consistently separate by age classes, sexual dimorphism, pelage variation, and environmental context, patterns valuable for population demographic analysis that would otherwise require manual annotation.

Our evaluation of a long-tailed distribution of species demonstrates that optimised HDBSCAN configurations maintain V-measure above 0.92 across both balanced and extremely imbalanced data typical of field deployments. This robustness, combined with the interpretable behaviour of rare species appearing either as outliers for expert review or merging with visually similar taxa, makes the approach practical for real-world camera trap workflows.

Our evaluation focused on mammals and birds from camera trap imagery, and performance may differ for taxonomic groups with lower morphological distinctiveness or imaging modalities. Additionally, the main benchmark uses 2D reduction throughout, and higher target dimensions were tested for DINOv3 only (Appendix E).

Future work should extend evaluation to additional taxonomic groups including fishes, insects, and reptiles where detection models already exist. Hierarchical clustering workflows that first separate species and then systematically reveal intra-specific structure would provide ecologists with multi-resolution views of their data.

Zero-shot clustering fundamentally transforms the annotation task from exhaustive image-by-image labelling to targeted review of cluster compositions. As camera trap networks expand globally and autonomous monitoring systems generate unprecedented image volumes, methods that organise unlabelled data before expert intervention become essential infrastructure for timely conservation decision-making. Our open-source toolkit, validated dataset, and comprehensive benchmarking results provide a foundation for continued development of zero-shot approaches in biodiversity monitoring.

### Author Contributions

**Hugo Markoff:** conceptualization, data curation, formal analysis, investigation, methodology, software, validation, visualization, writing – original draft, writing – review and editing. **Stefan Hein Bengtson:** conceptualization, methodology, supervision, writing – review and editing. **Michael Ørsted:** conceptualization, funding acquisition, methodology, project administration, supervision, validation, writing – review and editing.

### Funding

This work was carried out as part of the AI:EcoNet Lab and financially supported by the AI:X Labs initiative at Aalborg University, as well as by the Novo Nordisk Foundation (Grant NNF23OC0082599 to M.Ø.) and the Independent Research Fund Denmark (Grant 4251-00089B to M.Ø.).

### Conflicts of Interest

The authors declare no conflicts of interest.

### Data Availability Statement

The data that support the findings of this study are openly available in Hugging Face Datasets at https://huggingface.co/datasets/AI-EcoNet/ViT_animal_data. The benchmarking code is available at https://github.com/AI-EcoNet/zeroshot_clustering_animal_images, and all experiment logs and configurations at https://huggingface.co/datasets/AI-EcoNet/HUGO-Bench-Paper-Reproducibility. An overview is provided at https://hugomarkoff.github.io/animal_visual_transformer/. Recommended data citation: Markoff H; Bengtson S.H; Ørsted M; 2026; ViT animal data: Expert-validated camera-trap cropped detections for zero-shot clustering; Hugging Face Datasets; https://huggingface.co/datasets/AI-EcoNet/ViT_animal_data.

## Appendix A

### Evaluation Metrics Calculations

This appendix provides the complete mathematical formulations and explanations for the clustering evaluation metrics used in this study.

V-measure balances homogeneity and completeness through their harmonic mean, preventing optimization shortcuts where methods achieve high homogeneity by over-splitting or high completeness by collapsing all data into few clusters.

AMI quantifies information-theoretic agreement between predicted and ground truth partitions while correcting for chance agreement. Unlike raw mutual information or V-measure, AMI accounts for the fact that random clusterings achieve non-zero scores when compared to any partition. This adjustment is critical when comparing methods with vastly different cluster counts, as it prevents bias towards higher $K$ values.

AMI serves as a secondary validation metric, particularly for unsupervised clustering methods where automatic cluster count determination may deviate substantially from $S = 30$.

### Mathematical formulations

Let $\mathcal{C} = \{C_1, ..., C_K\}$ denote predicted clusters and $\mathcal{G} = \{G_1, ..., G_S\}$ denote ground truth species labels for $N$ images. Define:

- $n_{ks} = |C_k \cap G_s|$: number of images in cluster $k$ belonging to species $s$
- $n_k = |C_k| = \sum_s n_{ks}$: size of cluster $k$
- $n_s = |G_s| = \sum_k n_{ks}$: number of images of species $s$

*Homogeneity*

Homogeneity measures whether each cluster contains only members of a single species. It equals 1 when all clusters are 'pure' (each contains exactly one species) and approaches 0 when clusters contain uniform mixtures of all species.

$$h = \begin{cases} 1 & \text{if } H(\mathcal{G}) = 0 \\ 1 - \dfrac{H(\mathcal{G}|\mathcal{C})}{H(\mathcal{G})} & \text{otherwise} \end{cases}$$

where entropy of ground truth labels is:

$$H(\mathcal{G}) = -\sum_{s=1}^{S} \frac{n_s}{N} \log\left(\frac{n_s}{N}\right)$$

and conditional entropy of ground truth given cluster assignments is:

$$H(\mathcal{G}|\mathcal{C}) = -\sum_{k=1}^{K} \sum_{s=1}^{S} \frac{n_{ks}}{N} \log\left(\frac{n_{ks}}{n_k}\right)$$

Ecological interpretation: High homogeneity means ecologists can confidently assign species labels to clusters with minimal validation. Low homogeneity indicates clusters contain multiple species, requiring more exhaustive manual review to avoid mislabelling.

### Completeness

Completeness measures whether all members of a species are assigned to the same cluster. It equals 1 when each species occupies exactly one cluster and approaches 0 when every species is uniformly split across all clusters.

$$c = \begin{cases} 1 & \text{if } H(\mathcal{C}) = 0 \\ 1 - \dfrac{H(\mathcal{C}|\mathcal{G})}{H(\mathcal{C})} & \text{otherwise} \end{cases}$$

where entropy of cluster assignments is:

$$H(\mathcal{C}) = -\sum_{k=1}^{K} \frac{n_k}{N} \log\left(\frac{n_k}{N}\right)$$

and conditional entropy of clusters given ground truth is:

$$H(\mathcal{C}|\mathcal{G}) = -\sum_{s=1}^{S} \sum_{k=1}^{K} \frac{n_{ks}}{N} \log\left(\frac{n_{ks}}{n_s}\right)$$

Ecological interpretation: Low completeness means species images are scattered across multiple clusters, increasing validation burden. However, splits may reveal biologically meaningful subcategories (male/female dimorphism, juvenile/adult, seasonal pelage differences). Post-hoc inspection distinguishes informative fragmentation from pure failure.

### V-measure

V-measure is the harmonic mean of homogeneity and completeness, providing a balanced assessment:

$$V = 2 \cdot \frac{h \cdot c}{h + c}$$

The harmonic mean heavily penalises imbalance. A method achieving $h = 0.95$, $c = 0.50$ yields only $V = 0.655$, not the arithmetic mean of 0.725. This prevents extreme over-splitting (high $h$, low $c$) or under-splitting (low $h$, high $c$).

Range: $V \in [0, 1]$, where $V = 1$ requires perfect clustering ($h = c = 1$).

### Adjusted Mutual Information (AMI)

Mutual information quantifies shared information between clusterings:

$$\text{MI}(\mathcal{C}, \mathcal{G}) = \sum_{k=1}^{K} \sum_{s=1}^{S} \frac{n_{ks}}{N} \log\left(\frac{N \cdot n_{ks}}{n_k \cdot n_s}\right)$$

However, random clusterings achieve $\text{MI} > 0$ due to chance overlap. AMI corrects for this by subtracting expected MI under random permutation:

$$\text{AMI}(\mathcal{C}, \mathcal{G}) = \frac{\text{MI}(\mathcal{C}, \mathcal{G}) - \mathbb{E}[\text{MI}]}{\text{avg}(H(\mathcal{C}), H(\mathcal{G})) - \mathbb{E}[\text{MI}]}$$

Range: $\text{AMI} \in [0, 1]$ for similar cluster and species counts. Can be negative when clustering performs worse than random assignment.

Raw MI increases with cluster count, treating more clusters as more information, even if random. AMI corrects this bias.

When AMI and V-measure disagree: AMI is more sensitive to cluster count deviations. A method producing 90 clusters ($3\times$ true count) may achieve high V-measure if splits occur within species (high homogeneity maintained), but lower AMI due to information redundancy. In our experiments, AMI and V-measure correlated strongly ($r > 0.92$) for well-performing methods but diverged for pathological cases (DBSCAN with 200+ clusters), confirming their complementary roles.

## Appendix B

### Complete Dataset Composition

This appendix provides comprehensive details for all 60 species used in the study, including total image counts, validation status, and source datasets from LILA BC.

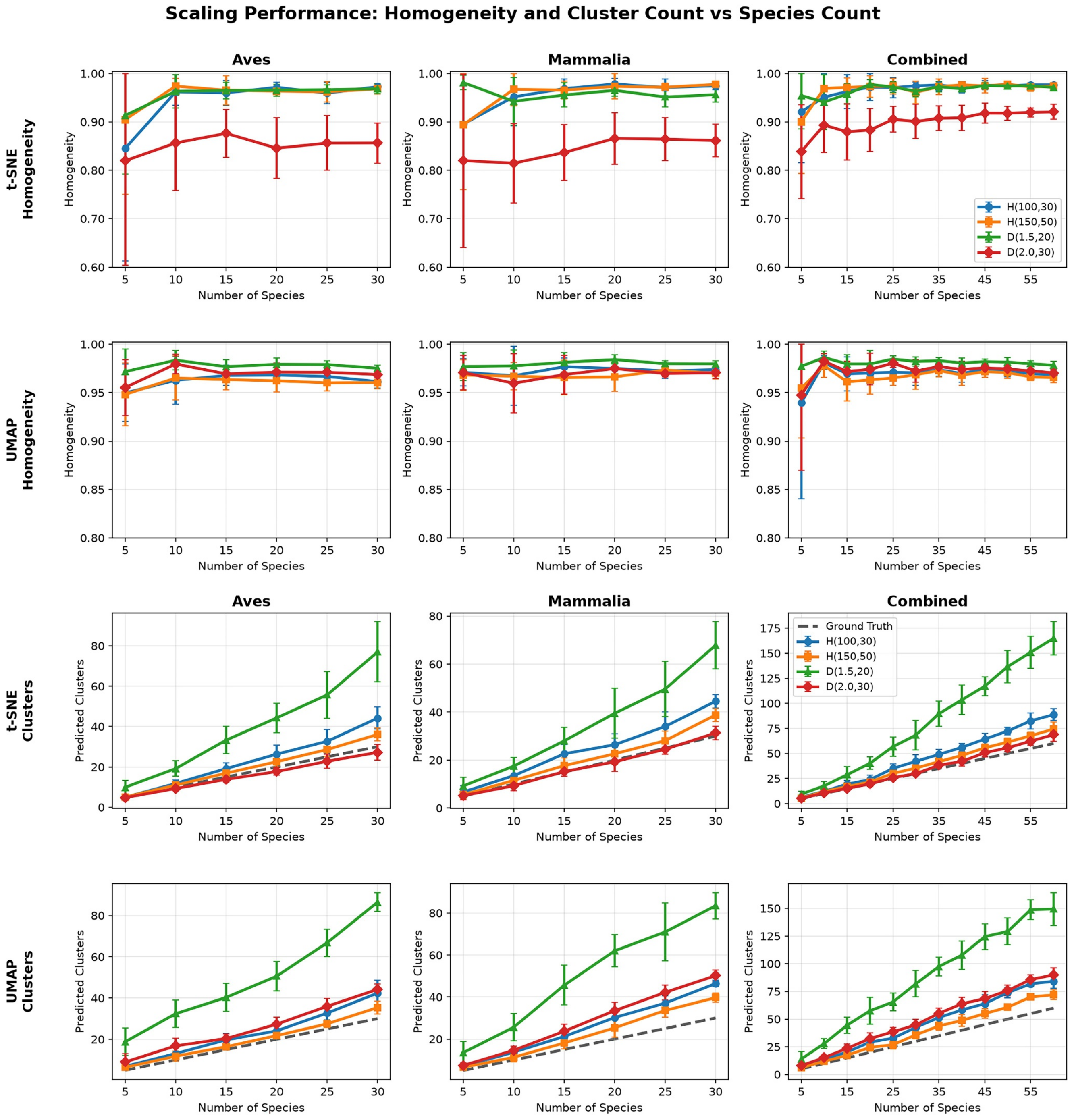

**FIGURE A1** | Scaling behaviour of optimised clustering configurations across varying species counts (5–30 for single-class, 5–60 for combined). Columns show Aves, Mammalia and the combined data. The upper two rows show homogeneity with t-SNE and UMAP dimensionality reduction, and the lower two rows the predicted cluster count with t-SNE and UMAP. Points show the mean over 10 runs and error bars ±1 SD, capped at the maximum homogeneity of 1. Dashed diagonal lines indicate ideal cluster count prediction. H = HDBSCAN configurations, D = DBSCAN configurations. See Table A6 for detailed performance metrics at 30/60 species.

**TABLE A1** | Complete species dataset composition (see Table 1 for source abbreviations).

| # | Class | Species name | Sources | Total | Valid | Uncertain |
|---|---|---|---|---|---|---|
| Aves (birds)—30 Species | | | | | | |
| 1 | Aves | American crow (*Corvus brachyrhynchos*) | ICT, SIC, ENA | 1240 | 1234 | 6 |
| 2 | Aves | Australasian swamphen (*Porphyrio melanotus*) | NZC | 2772 | 2772 | 0 |
| 3 | Aves | Australian magpie (*Gymnorhina tibicen*) | ICT, NZC | 2739 | 2733 | 6 |
| 4 | Aves | Black curassow (*Crax alector*) | WCS | 1949 | 1939 | 10 |
| 5 | Aves | Blue whistling thrush (*Myophonus caeruleus*) | SWG, WCS | 2157 | 2128 | 29 |
| 6 | Aves | Brown quail (*Synoicus ypsilophorus*) | NZC | 2662 | 2652 | 10 |
| 7 | Aves | Chicken (*Gallus gallus domesticus*) | ISC, SIC, NZC, ENA | 1647 | 1629 | 18 |
| 8 | Aves | Common chaffinch (*Fringilla coelebs*) | NZC | 2384 | 2370 | 14 |
| 9 | Aves | Common myna (*Acridotheres tristis*) | NZC | 2707 | 2704 | 3 |
| 10 | Aves | Dunnock (*Prunella modularis*) | NZC | 2066 | 2055 | 11 |
| 11 | Aves | European starling (*Sturnus vulgaris*) | NZC | 1876 | 1867 | 9 |
| 12 | Aves | Fantails (*Rhipidura*) | NZC | 2127 | 2121 | 6 |
| 13 | Aves | Greenfinch (*Chloris chloris*) | NZC | 2503 | 2483 | 20 |
| 14 | Aves | Kea (*Nestor notabilis*) | NZC | 2833 | 2814 | 19 |
| 15 | Aves | Kiwi (*Apteryx*) | NZC | 1634 | 1620 | 14 |
| 16 | Aves | Kori bustard (*Ardeotis kori*) | SS24, SCD, SKA, SKG, SSE, WCS | 1485 | 1484 | 1 |
| 17 | Aves | Mountain quail (*Oreortyx pictus*) | NAC, NZC | 2381 | 2337 | 44 |
| 18 | Aves | New Zealand robin (*Petroica australis*) | SIC, NZC | 2753 | 2719 | 34 |
| 19 | Aves | Ostrich (*Struthionidae*) | SS24, SCD, SKA, SKG, SSE, SMZ | 3727 | 3722 | 5 |
| 20 | Aves | Petrel (*Procellariidae*) | ISC | 2370 | 2354 | 16 |
| 21 | Aves | Pipit (*Anthus*) | NZC | 2316 | 2238 | 78 |
| 22 | Aves | Red junglefowl (*Gallus gallus*) | SWG, WCS | 2684 | 2671 | 13 |
| 23 | Aves | Spix's guan (*Penelope jacquacu*) | WCS | 3032 | 3022 | 10 |
| 24 | Aves | Swamp harrier (*Circus approximans*) | NZC | 2917 | 2917 | 0 |
| 25 | Aves | Takahē (*Porphyrio mantelli*) | NZC | 2085 | 2073 | 12 |
| 26 | Aves | Tū (*Prosthemadera novaeseelandiae*) | NZC | 2767 | 2747 | 20 |
| 27 | Aves | Vulturine guineafowl (*Acryllium vulturinum*) | WCS | 4420 | 4412 | 8 |
| 28 | Aves | Weka (*Gallirallus australis*) | NZC | 2421 | 2411 | 10 |
| 29 | Aves | Wild Turkey (*Meleagris gallopavo*) | NAC, ENA | 2057 | 2057 | 0 |
| 30 | Aves | Yellow-eyed penguin (*Megadyptes antipodes*) | NZC | 2817 | 2815 | 2 |
| Aves Subtotal | | | | 72,528 | 72,070 | 458 |
| Mammalia (Mammals)—30 Species | | | | | | |
| 31 | Mammals | Alpaca (*Vicugna pacos*) | WCS | 1864 | 1864 | 0 |
| 32 | Mammals | American black bear (*Ursus americanus*) | NAC, ENA, SIC, ICT | 2912 | 2797 | 115 |
| 33 | Mammals | Black rhinoceros (*Diceros bicornis*) | SMZ, DLC, WCS, SKA, SS24 | 1428 | 1419 | 9 |
| 34 | Mammals | Black-backed jackal (*Lupulella mesomelas*) | DLC, SMZ, SKG, SSE, SS24, SCD, SKA, WCS | 2067 | 2053 | 14 |
| 35 | Mammals | Bushpig (*Potamochoerus larvatus*) | SCD, WCS, SMZ, SS24, NKC | 1241 | 1230 | 11 |
| 36 | Mammals | Common brushtail possum (*Trichosurus vulpecula*) | WLC, UNS, NZC | 1365 | 1358 | 7 |
| 37 | Mammals | Common raccoon (*Procyon lotor*) | SIC, WCS, NAC, OSU, CAL | 2700 | 2577 | 123 |
| 38 | Mammals | Crab-eating mongoose (*Urva urva*) | SWG, WCS | 2364 | 2316 | 48 |
| 39 | Mammals | Crested porcupine (*Hystrix cristata*) | WCS, SKR, SSE, SCD, ICT, SKA, SMZ, SKG, SS24 | 1449 | 1449 | 0 |
| 40 | Mammals | Dromedary camel (*Camelus dromedarius*) | WCS | 1809 | 1798 | 11 |
| 41 | Mammals | Eastern gray squirrel (*Sciurus carolinensis*) | NAC, ENA | 2434 | 2414 | 20 |
| 42 | Mammals | Ferret badger (*Melogale*) | SWG | 1743 | 1671 | 72 |

(Continues)

**TABLE A1** | (Continued)

| # | Class | Species name | Sources | Total | Valid | Uncertain |
|---|---|---|---|---|---|---|
| 43 | Mammals | Gemsbok (*Oryx gazella*) | SCD, SKA, SS24, SMZ, SKG, DLC | 3963 | 3944 | 19 |
| 44 | Mammals | Giant armadillo (*Priodontes maximus*) | WCS | 463 | 461 | 2 |
| 45 | Mammals | Giraffe (*Giraffa camelopardalis*) | SS24, SSE, DLC, SKR, WCS | 2985 | 2974 | 11 |
| 46 | Mammals | Greater kudu (*Tragelaphus strepsiceros*) | DLC, WCS, NKC | 4410 | 4375 | 35 |
| 47 | Mammals | Hippopotamus (*Hippopotamus amphibius*) | SKR, WCS, SSE, SS24 | 2319 | 2314 | 5 |
| 48 | Mammals | Jaguar (*Panthera onca*) | WCS | 6855 | 6831 | 24 |
| 49 | Mammals | L'Hoest's monkey (*Allochrocebus lhoesti*) | WCS | 2941 | 2921 | 20 |
| 50 | Mammals | Least weasel (*Mustela nivalis*) | NZC | 1869 | 1869 | 0 |
| 51 | Mammals | New Zealand sea lion (*Phocarctos hookeri*) | NZC | 2296 | 2277 | 19 |
| 52 | Mammals | Northern treeshrew (*Tupaia belangeri*) | SWG, WCS | 699 | 687 | 12 |
| 53 | Mammals | Serval (*Leptailurus serval*) | SSE, SS24, WCS, NKC, SKR | 625 | 625 | 0 |
| 54 | Mammals | Ship rat (*Rattus rattus*) | WLC | 822 | 822 | 0 |
| 55 | Mammals | Spotted paca (*Cuniculus paca*) | WCS | 1715 | 1673 | 42 |
| 56 | Mammals | Stump-tailed macaque (*Macaca arctoides*) | SWG, WCS | 3558 | 3535 | 23 |
| 57 | Mammals | Sun bear (*Helarctos malayanus*) | WCS, SWG | 713 | 709 | 4 |
| 58 | Mammals | Warthog (*Phacochoerus africanus*) | SSE, SMZ, SS24, SKR | 1671 | 1666 | 5 |
| 59 | Mammals | White-nosed coati (*Nasua narica*) | WCS, SEN, MIS | 1996 | 1988 | 8 |
| 60 | Mammals | Wolf (*Canis lupus*) | ICT | 2307 | 2307 | 0 |
| Mammalia Subtotal | | | | 66,583 | 65,954 | 629 |
| GRAND TOTAL (60 Species) | | | | 139,111 | 138,024 | 1087 |

## Appendix C

### Supplementary Results

This appendix contains complete performance matrices for all experimental configurations tested in this study.

**TABLE A2** | Complete V-measure scores: All model × dimension reduction × clustering combinations—birds & mammals.

| | | Dim. reduction (avg. all clusterings) | | | | | Supervised $K = 30$ (UMAP + t-SNE) | | Unsupervised (UMAP + t-SNE) | |
|---|---|---|---|---|---|---|---|---|---|---|
| **Model** | **Class** | **UMAP** | **t-SNE** | **PCA** | **Isomap** | **KPCA** | **GMM** | **Hier.** | **HDBSCAN** | **DBSCAN** |
| DINOv3 | Birds | **0.877** | **0.884** | 0.456 | **0.686** | **0.527** | **0.957** | **0.960** | **0.932** | 0.671 |
| | Mammals | **0.869** | **0.879** | 0.458 | **0.642** | 0.295 | **0.952** | **0.952** | **0.914** | **0.677** |
| DINOv2 | Birds | 0.843 | 0.846 | **0.492** | 0.622 | 0.460 | 0.900 | 0.901 | 0.896 | **0.679** |
| | Mammals | 0.795 | 0.806 | **0.488** | 0.550 | **0.443** | 0.849 | 0.852 | 0.823 | **0.678** |
| BioCLIP 2 | Birds | 0.732 | 0.737 | 0.269 | 0.396 | 0.279 | 0.765 | 0.767 | 0.745 | 0.662 |
| | Mammals | 0.669 | 0.684 | 0.212 | 0.357 | 0.246 | 0.698 | 0.701 | 0.680 | 0.627 |
| CLIP | Birds | 0.680 | 0.689 | 0.202 | 0.294 | 0.213 | 0.705 | 0.707 | 0.692 | 0.633 |
| | Mammals | 0.642 | 0.661 | 0.197 | 0.276 | 0.196 | 0.660 | 0.663 | 0.666 | 0.618 |
| SigLIP | Birds | 0.654 | 0.663 | 0.215 | 0.272 | 0.218 | 0.675 | 0.680 | 0.665 | 0.615 |
| | Mammals | 0.625 | 0.640 | 0.242 | 0.298 | 0.205 | 0.637 | 0.641 | 0.646 | 0.605 |

*Note:* All values averaged across 100 runs (10 subsamples × 10 dimensionality reduction parameter configurations). Dim. Reduction columns: Average V-measure across all 4 clustering methods (GMM $K = 30$, Hier. $K = 30$, HDBSCAN, DBSCAN) Bold, best per class per metric.
Abbreviations: Hier., Hierarchical Clustering; KPCA, Kernel PCA. The bold marks the best value per taxonomic class for each method.

**TABLE A3** | Supervised clustering performance: Effect of cluster count (K) on V-measure and AMI.

| | | $K = 15$ | | $K = 30$ | | $K = 45$ | | $K = 90$ | | $K = 180$ | | Best |
|---|---|---|---|---|---|---|---|---|---|---|---|---|
| **Model** | **Method** | **V-M** | **AMI** | **V-M** | **AMI** | **V-M** | **AMI** | **V-M** | **AMI** | **V-M** | **AMI** | ***K*** |
| DINOv3 | Hierarchical | 0.801 | 0.788 | **0.909** | **0.908** | 0.880 | 0.878 | 0.810 | 0.807 | 0.753 | 0.749 | 30 |
| DINOv3 | GMM | 0.793 | 0.780 | **0.909** | **0.907** | 0.880 | 0.877 | 0.811 | 0.808 | 0.754 | 0.750 | 30 |
| DINOv2 | Hierarchical | 0.740 | 0.726 | **0.838** | **0.833** | 0.825 | 0.820 | 0.769 | 0.762 | 0.721 | 0.713 | 30 |
| DINOv2 | GMM | 0.731 | 0.717 | **0.837** | **0.832** | 0.823 | 0.818 | 0.768 | 0.761 | 0.722 | 0.714 | 30 |
| BioCLIP 2 | Hierarchical | 0.608 | 0.588 | 0.686 | 0.673 | **0.697** | **0.685** | 0.668 | 0.655 | 0.641 | 0.627 | 45 |
| BioCLIP 2 | GMM | 0.599 | 0.579 | 0.685 | 0.672 | **0.693** | **0.681** | 0.665 | 0.652 | 0.640 | 0.626 | 45 |
| CLIP | Hierarchical | 0.563 | 0.540 | 0.634 | 0.617 | **0.652** | **0.637** | 0.639 | 0.623 | 0.619 | 0.603 | 45 |
| CLIP | GMM | 0.556 | 0.533 | 0.632 | 0.615 | **0.648** | **0.633** | 0.636 | 0.620 | 0.618 | 0.602 | 45 |
| SigLIP | Hierarchical | 0.542 | 0.519 | 0.614 | 0.596 | **0.633** | **0.617** | 0.623 | 0.606 | 0.606 | 0.589 | 45 |
| SigLIP | GMM | 0.535 | 0.512 | 0.611 | 0.593 | **0.628** | **0.612** | 0.619 | 0.602 | 0.604 | 0.587 | 45 |

*Note:* Ground Truth: 30 species. Averages across 10 runs with UMAP and t-SNE combined.
Abbreviations: AMI = Adjusted Mutual Information; V-M = V-measure.

**TABLE A4** | Unsupervised clustering: Cluster count prediction (Ground truth: 30 species)—Birds versus Mammals.

| | | | Birds (Aves) | | | Mammals (Mammalia) | | |
|---|---|---|---|---|---|---|---|---|
| **Model** | **Dim. red.** | **Method** | **Pred. clust.** | **Range** | **v-Meas.** | **Pred. clust.** | **Range** | **V-meas.** |
| **DINOv3** | t-SNE | HDBSCAN | **33.2 ± 2.5** | (29–40) | **0.948** | **37.4 ± 3.0** | (29–43) | **0.939** |
| DINOv3 | t-SNE | DBSCAN | 255.5 ± 13.1 | (224–278) | 0.668 | 247.2 ± 10.8 | (211–270) | 0.668 |
| DINOv3 | UMAP | HDBSCAN | 47.3 ± 5.0 | (34–56) | 0.917 | 57.1 ± 4.4 | (44–71) | 0.890 |
| DINOv3 | UMAP | DBSCAN | 204.0 ± 12.3 | (179–236) | 0.674 | 189.4 ± 9.3 | (173–212) | 0.687 |
| **DINOv2** | t-SNE | HDBSCAN | 33.8 ± 2.4 | (30–39) | 0.903 | 44.9 ± 4.3 | (38–56) | 0.843 |
| DINOv2 | t-SNE | DBSCAN | 220.6 ± 11.9 | (199–247) | 0.677 | 217.9 ± 12.5 | (198–246) | 0.672 |
| DINOv2 | UMAP | HDBSCAN | 38.9 ± 2.6 | (32–46) | 0.889 | 54.5 ± 4.4 | (43–65) | 0.802 |
| DINOv2 | UMAP | DBSCAN | 172.7 ± 10.8 | (146–198) | 0.680 | 149.0 ± 11.2 | (124–176) | 0.685 |
| **BioCLIP 2** | t-SNE | HDBSCAN | 57.1 ± 5.1 | (46–69) | 0.760 | 59.2 ± 4.5 | (45–66) | 0.700 |
| BioCLIP 2 | t-SNE | DBSCAN | 220.3 ± 9.5 | (201–239) | 0.652 | 225.8 ± 8.5 | (204–246) | 0.630 |
| BioCLIP 2 | UMAP | HDBSCAN | 64.3 ± 6.7 | (43–80) | 0.729 | 70.3 ± 6.7 | (55–86) | 0.659 |
| BioCLIP 2 | UMAP | DBSCAN | 136.8 ± 8.7 | (112–158) | 0.672 | 150.4 ± 7.8 | (129–175) | 0.625 |
| **CLIP** | t-SNE | HDBSCAN | 59.1 ± 5.3 | (48–69) | 0.708 | 69.5 ± 5.4 | (58–79) | 0.686 |
| CLIP | t-SNE | DBSCAN | 244.3 ± 7.8 | (228–265) | 0.630 | 240.7 ± 10.6 | (219–265) | 0.627 |
| CLIP | UMAP | HDBSCAN | 67.5 ± 5.3 | (51–80) | 0.677 | 76.2 ± 6.6 | (54–90) | 0.646 |
| CLIP | UMAP | DBSCAN | 140.9 ± 9.3 | (116–162) | 0.637 | 162.0 ± 9.6 | (135–184) | 0.610 |
| **SigLIP** | t-SNE | HDBSCAN | 61.8 ± 5.1 | (49–71) | 0.681 | 70.2 ± 6.8 | (50–83) | 0.661 |
| SigLIP | t-SNE | DBSCAN | 239.3 ± 7.9 | (226–254) | 0.619 | 248.9 ± 9.8 | (228–267) | 0.612 |
| SigLIP | UMAP | HDBSCAN | 68.2 ± 7.0 | (53–87) | 0.648 | 77.3 ± 6.9 | (46–89) | 0.631 |
| SigLIP | UMAP | DBSCAN | 154.5 ± 9.5 | (135–192) | 0.611 | 166.9 ± 10.1 | (140–193) | 0.598 |

*Note:* Predicted clusters shown as Mean ± SD. Best cluster prediction (closest to 30) and V-measure per class shown in bold. All values averaged across 100 runs (10 subsamples × 10 dimension reduction runs).

**TABLE A5** | Complete clustering metrics for top-performing configurations at $K = 30$: Birds versus Mammals.

| | | | Birds (Aves) | | | | | | Mammals (Mammalia) | | | | | |
|---|---|---|---|---|---|---|---|---|---|---|---|---|---|---|
| **Model** | **Dim. red.** | **Clustering** | **V-M** | **AMI** | **ARI** | **Hom.** | **Cmp.** | **Sil.** | **V-M** | **AMI** | **ARI** | **Hom.** | **Cmp.** | **Sil.** |
| **DINOv3** | t-SNE | Hierarchical | **0.961** | **0.960** | **0.943** | **0.961** | **0.961** | 0.714 | **0.954** | **0.953** | **0.919** | **0.953** | 0.956 | 0.685 |
| DINOv3 | t-SNE | GMM | 0.958 | 0.957 | 0.927 | 0.956 | 0.960 | 0.708 | 0.954 | 0.953 | 0.914 | 0.951 | **0.956** | 0.679 |
| DINOv3 | UMAP | Hierarchical | 0.960 | 0.959 | 0.941 | 0.959 | 0.960 | 0.868 | 0.949 | 0.948 | 0.907 | 0.947 | 0.951 | 0.821 |
| DINOv3 | UMAP | GMM | 0.956 | 0.955 | 0.925 | 0.954 | 0.958 | **0.872** | 0.949 | 0.948 | 0.904 | 0.947 | 0.952 | **0.823** |
| **DINOv2** | t-SNE | Hierarchical | 0.903 | 0.900 | 0.834 | 0.900 | 0.905 | 0.711 | 0.855 | 0.852 | 0.742 | 0.851 | 0.860 | 0.630 |
| DINOv2 | t-SNE | GMM | 0.900 | 0.898 | 0.827 | 0.897 | 0.903 | 0.700 | 0.853 | 0.850 | 0.723 | 0.848 | 0.859 | 0.618 |

(Continues)

**TABLE A5** | (Continued)

| Model | Dim. red. | Clustering | Birds (Aves) | | | | | | Mammals (Mammalia) | | | | | |
|---|---|---|---|---|---|---|---|---|---|---|---|---|---|---|
| | | | V-M | AMI | ARI | Hom. | Cmp. | Sil. | V-M | AMI | ARI | Hom. | Cmp. | Sil. |
| DINOv2 | UMAP | Hierarchical | 0.900 | 0.898 | 0.829 | 0.898 | 0.903 | 0.857 | 0.848 | 0.845 | 0.719 | 0.843 | 0.854 | 0.744 |
| DINOv2 | UMAP | GMM | 0.901 | 0.898 | 0.829 | 0.898 | 0.903 | 0.851 | 0.845 | 0.842 | 0.702 | 0.839 | 0.852 | 0.730 |
| **BioCLIP 2** | t-SNE | Hierarchical | 0.770 | 0.765 | 0.606 | 0.765 | 0.776 | 0.558 | 0.703 | 0.696 | 0.512 | 0.698 | 0.708 | 0.502 |
| BioCLIP 2 | t-SNE | GMM | 0.767 | 0.762 | 0.588 | 0.761 | 0.774 | 0.554 | 0.703 | 0.696 | 0.506 | 0.696 | 0.709 | 0.496 |
| BioCLIP 2 | UMAP | Hierarchical | 0.764 | 0.758 | 0.572 | 0.756 | 0.772 | 0.672 | 0.699 | 0.693 | 0.491 | 0.691 | 0.708 | 0.586 |
| BioCLIP 2 | UMAP | GMM | 0.763 | 0.758 | 0.564 | 0.754 | 0.772 | 0.658 | 0.693 | 0.687 | 0.473 | 0.683 | 0.704 | 0.557 |
| **CLIP** | t-SNE | Hierarchical | 0.708 | 0.701 | 0.528 | 0.704 | 0.712 | 0.491 | 0.668 | 0.661 | 0.460 | 0.663 | 0.674 | 0.460 |
| CLIP | t-SNE | GMM | 0.709 | 0.702 | 0.523 | 0.704 | 0.714 | 0.487 | 0.665 | 0.657 | 0.450 | 0.658 | 0.671 | 0.451 |
| CLIP | UMAP | Hierarchical | 0.706 | 0.700 | 0.513 | 0.701 | 0.712 | 0.602 | 0.658 | 0.651 | 0.432 | 0.649 | 0.668 | 0.556 |
| CLIP | UMAP | GMM | 0.701 | 0.695 | 0.491 | 0.694 | 0.709 | 0.576 | 0.656 | 0.648 | 0.415 | 0.644 | 0.668 | 0.537 |
| **SigLIP** | t-SNE | Hierarchical | 0.678 | 0.671 | 0.479 | 0.673 | 0.683 | 0.466 | 0.644 | 0.636 | 0.439 | 0.639 | 0.648 | 0.455 |
| SigLIP | t-SNE | GMM | 0.675 | 0.667 | 0.469 | 0.669 | 0.680 | 0.468 | 0.642 | 0.635 | 0.435 | 0.637 | 0.648 | 0.455 |
| SigLIP | UMAP | Hierarchical | 0.682 | 0.674 | 0.474 | 0.675 | 0.689 | 0.568 | 0.638 | 0.629 | 0.418 | 0.630 | 0.645 | 0.534 |
| SigLIP | UMAP | GMM | 0.675 | 0.668 | 0.450 | 0.666 | 0.684 | 0.549 | 0.631 | 0.623 | 0.392 | 0.621 | 0.642 | 0.503 |

*Note:* Birds consistently achieve higher clustering performance than mammals across all models (average V-measure difference of 0.042). All values averaged across 100 runs (10 subsamples × 10 dimension reduction runs). Best value per metric per class shown in bold.
Abbreviations: AMI = Adjusted Mutual Info, ARI = Adjusted Rand Index, Cmp. = Completeness, Hom. = Homogeneity, Metrics: V-M = V-measure, Sil. = Silhouette Score.

## Appendix D

### Extended Clustering Results for Uneven Distributions

This appendix provides comprehensive clustering results for the extreme uneven distribution scenario discussed in Section 4.8. We systematically evaluated 10 parameter configurations for both HDBSCAN and DBSCAN, testing each with both t-SNE and UMAP dimensionality reduction.

**TABLE A6** | Clustering performance on extreme uneven distributions (20–MAX images per species).

| Config | Aves (30 species) | | | | Mammalia (30 species) | | | | Combined (60 species) | | | |
|---|---|---|---|---|---|---|---|---|---|---|---|---|
| | V | H | Cl. | O% | V | H | Cl. | O% | V | H | Cl. | *N*% |
| *t-SNE* | | | | | | | | | | | | |
| HDBSCAN (15, 5) | 0.776 | 0.977 | 415 | 14.2 | 0.792 | 0.978 | 304 | 12.2 | 0.798 | 0.985 | 781 | 15.9 |
| HDBSCAN (30, 10) | 0.876 | 0.977 | 120 | 7.7 | 0.868 | 0.973 | 97 | 6.2 | 0.880 | 0.981 | 254 | 9.2 |
| HDBSCAN (50, 20) | 0.921 | 0.972 | 58 | 4.7 | 0.907 | 0.975 | 57 | 3.9 | 0.922 | 0.976 | 124 | 5.0 |
| HDBSCAN (75, 25) | 0.931 | 0.973 | 48 | 4.7 | 0.912 | 0.974 | 50 | 3.8 | 0.929 | 0.976 | 103 | 4.6 |
| HDBSCAN (100, 30) | 0.936 | 0.974 | 43 | 4.7 | 0.918 | 0.973 | 44 | 3.9 | 0.936 | 0.977 | 91 | 4.2 |
| HDBSCAN (150, 50) | **0.946** | 0.971 | 36 | 4.7 | **0.930** | 0.973 | 39 | 3.7 | **0.948** | 0.973 | 72 | 4.1 |
| DBSCAN (1.0, 10) | 0.694 | 0.990 | 699 | 21.3 | 0.700 | 0.989 | 575 | 21.1 | 0.734 | 0.991 | 1197 | 21.6 |
| DBSCAN (1.2, 15) | 0.813 | 0.983 | 226 | 8.2 | 0.817 | 0.982 | 176 | 8.2 | 0.830 | 0.985 | 436 | 9.1 |
| DBSCAN (1.5, 20) | 0.909 | 0.972 | 75 | 2.3 | 0.887 | 0.963 | 68 | 2.1 | 0.900 | 0.972 | 167 | 2.8 |
| DBSCAN (2.0, 30) | 0.911 | 0.872 | 28 | 0.3 | 0.899 | 0.877 | 32 | 0.2 | 0.930 | 0.922 | 68 | 0.5 |
| *UMAP* | | | | | | | | | | | | |
| HDBSCAN (15, 5) | 0.869 | 0.972 | 178 | 6.4 | 0.843 | 0.978 | 179 | 7.9 | 0.893 | 0.978 | 294 | 6.8 |
| HDBSCAN (30, 10) | 0.897 | 0.968 | 91 | 4.6 | 0.878 | 0.977 | 95 | 5.5 | 0.912 | 0.976 | 164 | 5.1 |
| HDBSCAN (50, 20) | 0.911 | 0.966 | 62 | 3.7 | 0.901 | 0.975 | 62 | 3.6 | 0.920 | 0.973 | 122 | 4.1 |
| HDBSCAN (75, 25) | 0.916 | 0.964 | 51 | 3.3 | 0.910 | 0.975 | 52 | 3.1 | 0.926 | 0.971 | 102 | 3.4 |
| HDBSCAN (100, 30) | 0.922 | 0.962 | 45 | 3.0 | 0.920 | 0.973 | 45 | 2.1 | 0.933 | 0.971 | 89 | 3.1 |
| HDBSCAN (150, 50) | 0.933 | 0.955 | 36 | 2.3 | **0.926** | 0.971 | 40 | 1.9 | **0.942** | 0.967 | 74 | 2.3 |
| DBSCAN (1.0, 10) | 0.763 | 0.984 | 382 | 19.3 | 0.753 | 0.987 | 347 | 21.9 | 0.814 | 0.986 | 608 | 21.9 |
| DBSCAN (1.2, 15) | 0.855 | 0.979 | 161 | 7.2 | 0.826 | 0.984 | 159 | 10.1 | 0.871 | 0.984 | 284 | 10.7 |

(Continues)

**TABLE A6** | (Continued)

| Config | Aves (30 species) | | | | Mammalia (30 species) | | | | Combined (60 species) | | | |
|---|---|---|---|---|---|---|---|---|---|---|---|---|
| | V | H | Cl. | O% | V | H | Cl. | O% | V | H | Cl. | *N*% |
| DBSCAN (1.5, 20) | 0.910 | 0.974 | 80 | 2.9 | 0.877 | 0.979 | 85 | 3.9 | 0.910 | 0.980 | 155 | 4.3 |
| DBSCAN (2.0,30) | **0.940** | 0.964 | 45 | 1.6 | 0.914 | 0.967 | 50 | 1.1 | 0.939 | 0.973 | 95 | 1.6 |

*Note:* Metrics: V = V-measure (harmonic mean of homogeneity and completeness), H = Homogeneity (each cluster contains only members of a single class), Cl. = Number of clusters found (true values: 30/30/60), O% = Percentage of points classified as outliers. Results averaged over 10 random samples. HDBSCAN (min_cluster_size, min_samples), DBSCAN ($\epsilon\times$ multiplier, min_samples). Bold values indicate best V-measure per dimensionality reduction method.

## Appendix E

### Effect of Target Dimensionality

Table A7 reports clustering of DINOv3 embeddings reduced to target dimensions from 2D to 256D, using HDBSCAN (15, 5) on the same 10 image subsets per taxonomic class as the main benchmark (one seed per subset, $n$ = 20 per cell). Because the configurations here reject very different fractions of images, noise points are scored as their own cluster rather than excluded, which gives lower values than in the main tables. t-SNE is shown only at 2D and 3D, the output dimensions supported by the Barnes-Hut implementation used in the main benchmark van der Maaten [56].

UMAP improves up to about 32D and then plateaus, with its predicted cluster count approaching the true value, and from 8D upwards it matches or slightly exceeds 2D t-SNE.

**TABLE A7** | Effect of target dimensionality on clustering of DINOv3 embeddings with HDBSCAN (15, 5).

| | 2D | 3D | 8D | 16D | 32D | 64D | 128D | 256D |
|---|---|---|---|---|---|---|---|---|
| V-measure | | | | | | | | |
| t-SNE | 0.937 | 0.928 | — | — | — | — | — | — |
| UMAP | 0.901 | 0.919 | 0.937 | 0.941 | 0.943 | 0.944 | 0.943 | 0.942 |
| Predicted clusters | | | | | | | | |
| t-SNE | 37.3 | 33.0 | — | — | — | — | — | — |
| UMAP | 52.7 | 45.2 | 37.0 | 34.7 | 33.0 | 32.9 | 33.0 | 33.8 |

*Note:* The upper block gives the V-measure with noise scored as its own cluster, and the lower block the mean number of predicted clusters (true count 30). Raw 1280D embeddings without reduction give a V-measure of 0.813. t-SNE is shown at its supported output dimensions only.

## Appendix F

### Compute Requirements

Table A8 reports the measured runtime and memory requirements of the full pipeline on a single consumer workstation.

**TABLE A8** | Measured compute cost of the pipeline on one consumer workstation with an NVIDIA GeForce RTX 5080 GPU (15.9 GB VRAM), an Intel Core Ultra 9285 CPU (24 cores) and 64 GB RAM, running PyTorch 2.10 with CUDA 12.8.

| Embedding extraction (GPU) | | | | |
|---|---|---|---|---|
| **Model** | ***s*** | **Peak VRAM** | **12,000 images** | **139,111 images** |
| DINOv3 | 65 | 3.7 GB | 3.1 min | 36 min |
| DINOv2 | 40 | 5.1 GB | 4.9 min | 57 min |
| BioCLIP 2 | 151 | 1.5 GB | 1.3 min | 15 min |
| CLIP | 134 | 1.7 GB | 1.5 min | 17 min |
| SigLIP | 592 | 0.6 GB | 0.3 min | 4 min |

| Reduction and clustering, 12,000 images (CPU) | |
|---|---|
| **Step** | **Time** |
| t-SNE to 2D | 109 s |
| UMAP to 2D | 57 s |
| HDBSCAN | 0.3 s |
| Hierarchical | 6 s |

*Note:* The upper block covers embedding extraction on the GPU (fp32, batch size 32 as in the main benchmark), which runs once per dataset and is reused for all analyses. Peak VRAM is the maximum GPU memory allocated during extraction, and the last two columns give the extraction time for 12,000 images and for the full dataset. The lower block covers dimensionality reduction to 2D and clustering on the CPU for 12,000 images, repeated for each analysis.